\documentclass[10pt,twocolumn,letterpaper]{article}

\usepackage[pagenumbers]{cvpr} %
\usepackage[table]{xcolor}

\usepackage{booktabs}
\usepackage{multirow}
\usepackage{xspace}
\usepackage{titletoc}
\usepackage{minitoc}
\definecolor{lightblue}{RGB}{222,239,255}
\definecolor{verylightgray}{gray}{0.93}
\definecolor{cvprblue}{rgb}{0.21,0.49,0.74}
\usepackage[pagebackref,breaklinks,colorlinks,citecolor=cvprblue]{hyperref}

\newcommand{\methodName}{$\infty$-RoPE\xspace}

\title{Infinity-RoPE: Action-Controllable Infinite Video Generation Emerges From Autoregressive Self-Rollout}
\author{Hidir Yesiltepe$^1$ \quad
  Tuna Han Salih Meral$^1$ \quad 
  Adil Kaan Akan$^2$ \quad
  Kaan Oktay$^2$ \quad
  Pinar Yanardag$^1$ \\[0.25em] 
  $^1$Virginia Tech \qquad $^2$fal \\ [0.5em]
\normalsize{Project Page: \url{https://infinity-rope.github.io}}}

\begin{document}
\twocolumn[{
\maketitle
\begin{center}
    \captionsetup{type=figure}
    \vspace{-1em}
\newcommand{\imwidth}{1\textwidth}
\begin{tabular}{@{}c@{}}
\parbox{\imwidth}{ \centering \includegraphics[width=\imwidth, ]{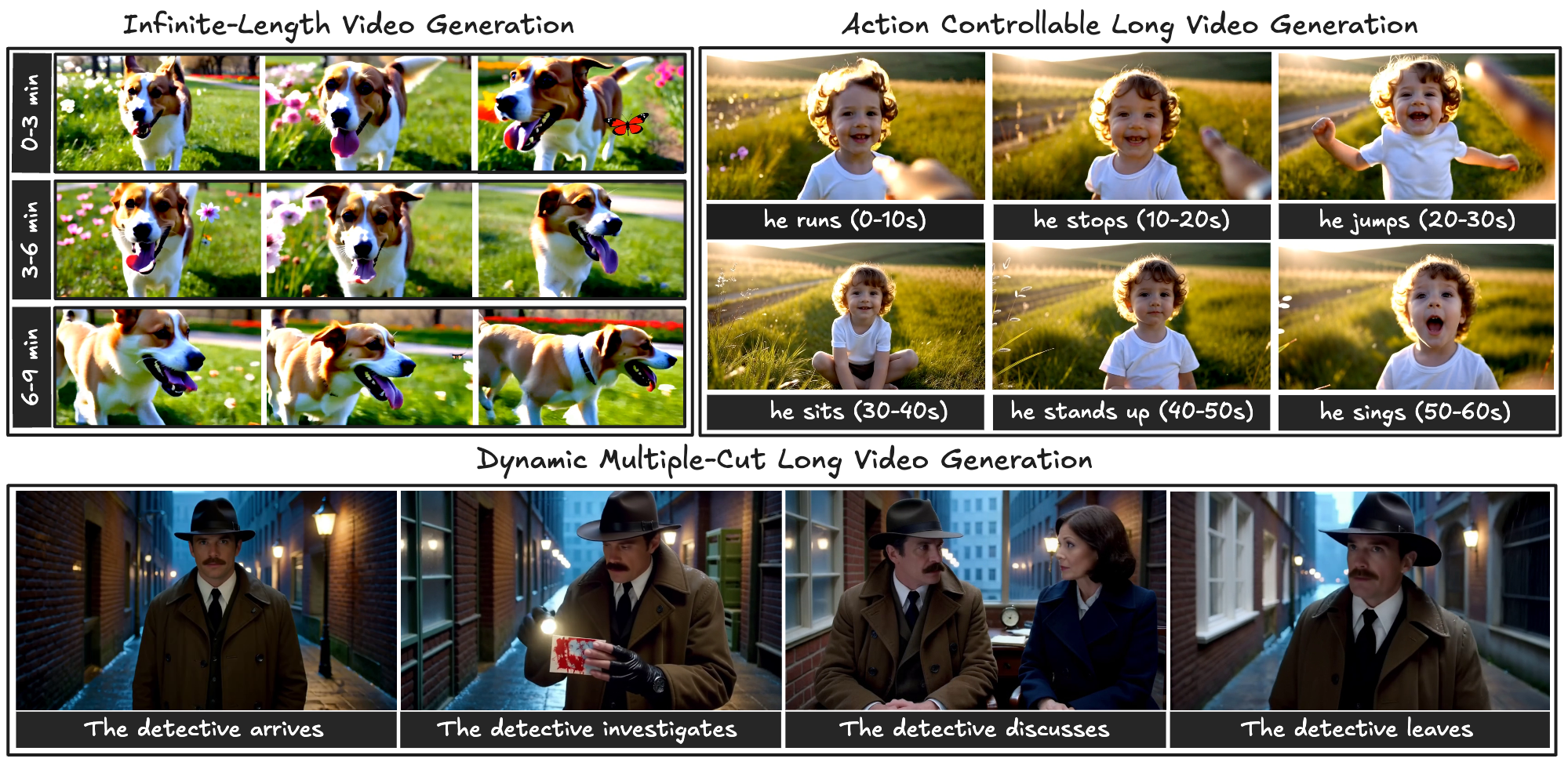}}
\\

\vspace{1em}
\end{tabular}
    \vspace{-2.7em}
    \captionof{figure}{\methodName demonstrates three core capabilities: Infinite-length video generation enabled by Block-Relativistic RoPE, fine-grained action-control through KV Flush, and cinematic multi-cut scene composition via RoPE Cut. }
    \label{fig:teaser}
\end{center}
}] 
    
\begin{abstract}
Current autoregressive video diffusion models are constrained by three core bottlenecks: (i) the finite temporal horizon imposed by the base model's 3D Rotary Positional Embedding (3D-RoPE), (ii) slow prompt responsiveness in maintaining fine-grained action control during long-form rollouts, and (iii) the inability to realize discontinuous cinematic transitions within a single generation stream. We introduce \methodName, a unified inference-time framework that addresses all three limitations through three interconnected components: \textit{Block-Relativistic RoPE}, \textit{KV Flush}, and \textit{RoPE Cut}. Block-Relativistic RoPE reformulates temporal encoding as a moving local reference frame, where each newly generated latent block is rotated relative to the base model's maximum frame horizon while earlier blocks are rotated backward to preserve relative temporal geometry. This relativistic formulation eliminates fixed temporal positions, enabling continuous video generation far beyond the base positional limits. To obtain fine-grained action control without re-encoding, \textit{KV Flush} renews the KV cache by retaining only two latent frames, the global sink and the last generated latent frame, thereby ensuring immediate prompt responsiveness. Finally, \textit{RoPE Cut} introduces controlled discontinuities in temporal RoPE coordinates, enabling multi-cut scene transitions within a single continuous rollout. Together, these components establish \methodName as a training-free foundation for infinite-horizon, controllable, and cinematic video diffusion. Comprehensive experiments show that \methodName consistently surpasses previous autoregressive models in overall VBench scores.

\end{abstract}
    
\section{Introduction}
\label{sec:intro}
In recent years, video diffusion models have advanced rapidly, progressing from early pixel-space denoisers~\cite{ho2020denoising,song2020denoising,ho2022video} to transformer-based architectures~\cite{ho2022imagen,singer2022make,blattmann2023align,blattmann2023stable}. The introduction of Diffusion Transformers (DiTs)~\cite{peebles2023scalable,gupta2024photorealistic} has further shifted the scaling frontier of generative video modeling, enabling high-resolution, multi-shot synthesis and significantly reducing the quality gap between synthetic and real-world video. Despite these advances, existing models are still limited to short video durations.

To overcome these shortcomings, recent research has shifted toward autoregressive video generation frameworks~\cite{weissenborn2019scaling, cui2025self, yin2025slow, huang2025self, liu2025rolling, kodaira2025streamdit, deng2024autoregressive, chen2025skyreels} that align with the causal nature of time. CausVid~\cite{yin2025slow} first demonstrated that a bidirectional DiT could be distilled into a causal student through Distribution Matching Distillation (DMD)~\cite{yin2024one, yin2024improved}, converting dense bidirectional attention into block-causal attention. However, its asymmetric teacher--student supervision introduced a train--test mismatch leading to exposure bias and degraded visual quality in long rollouts. Self-Forcing~\cite{huang2025self} addressed this issue by performing autoregressive self-rollout during training, using its own generated frames and a rolling KV cache to align training with inference and achieve temporally consistent short-form generation. Building upon this paradigm, Self-Forcing++~\cite{cui2025self} extended the framework to minute-scale horizons through long rollouts and extended DMD, adapting the temporal 3D RoPE ~\cite{su2024roformer} across the full 1024-frame dimension while still supervising only short slices. Rolling Forcing~\cite{liu2025rolling} further refined this direction by jointly denoising multiple consecutive frames with progressively increasing noise levels within a rolling window.

Despite these advances, prior autoregressive diffusion methods remain fundamentally constrained by the architectural limits of their positional strategies. Along the temporal dimension, 3D-RoPE restricts autoregressive diffusion transformers to a fixed horizon of 1024 latent frames, beyond which temporal encoding collapses and attention degenerates. At the same time, extending the Self-Forcing training paradigm to such horizons incurs prohibitively slow optimization and high computational cost due to sequential self-rollouts, as reported in~\cite{cui2025self}.

\begin{figure}[t]
    \centering
    \includegraphics[width=1.0\linewidth]{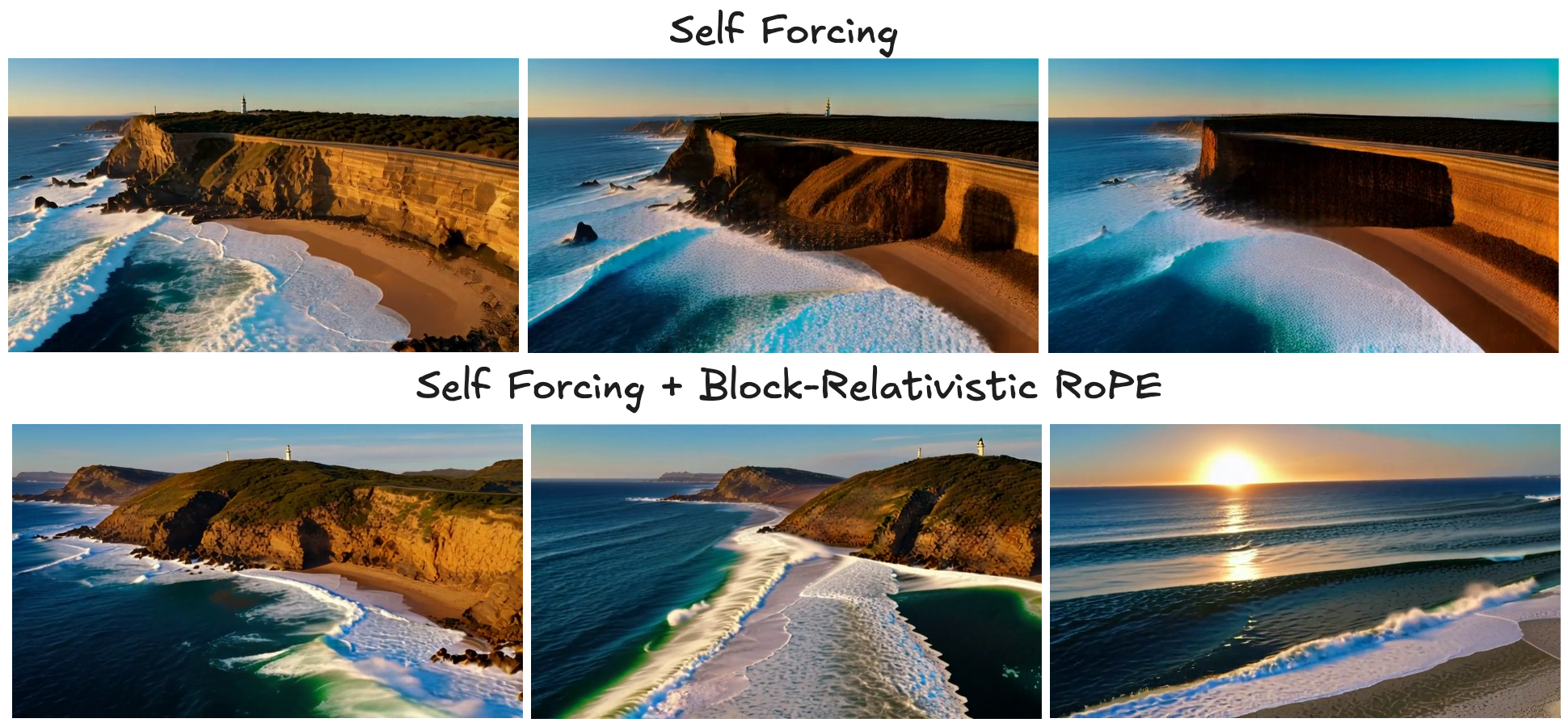} \\
    \caption{\textbf{Motivation.} Thirty-second video generation with Self-Forcing combined with our method \textbf{(Top)} Self-Forcing alone cannot sustain dynamic long-form generation. \textbf{(Bottom)} When augmented with Block-Relativistic RoPE, a Self-Forcing model trained only on five-second videos produces highly dynamic, high-quality long-form sequences.}
\label{fig:motivation}
\vspace{-1.5em}
\end{figure}

In this work, we revisit the challenge of long-form video generation from a fundamentally different standpoint. Prior autoregressive diffusion frameworks extend temporal scalability through longer self-rollouts~\cite{cui2025self, liu2025rolling} or retraining on long-video data~\cite{chen2025skyreels}, yet they are limited by the absolute frame indexing of 3D-RoPE and the heavy memory footprint of uncompressed KV caches. Instead, we investigate whether it is possible to overcome these bottlenecks purely through relativistic adaptation and training-free architectural reparameterization, without relying on any long-video supervision. We show that autoregressive video DiTs trained under the Self-Forcing paradigm for only 5-second clips already possess the capacity for highly dynamic infinite-horizon generation, as illustrated in Fig.~\ref{fig:motivation}. 

To address these challenges, we introduce Block-Relativistic RoPE, a training-free relativistic positional encoding that redefines temporal coordinates as a moving reference frame, allowing each newly generated block to be rotated relative to the base model’s maximum frame horizon while earlier blocks are rotated backward to preserve relative geometry. This formulation removes fixed temporal positions and enables continuous generation far beyond the 1024-frame RoPE limit. We further propose KV Flush, which leverages this relativistic geometry to renew the KV cache and resume generation from past temporal indices, providing prompt-responsive action control with constant memory. Finally, RoPE Cut performs controlled discontinuities in temporal coordinates to introduce cinematic multi-cut transitions within a single rollout. Altogether, these components transform short-horizon Self-Forcing models into infinite-horizon, scene-aware generators. Our contributions are summarized as follows:
\begin{itemize}
    \item We propose \textbf{\methodName}, a training-free methodology that converts existing short-horizon autoregressive self-rollout diffusion models into action-controllable infinite-horizon generators.
    \item We introduce \textbf{Block-Relativistic RoPE}, a relativistic positional encoding that reformulates temporal structure as a moving reference frame, enabling generation far beyond the 1024-frame RoPE limit.
    \item We develop two inference-time operators, \textbf{KV Flush} and \textbf{RoPE Cut}, which respectively provide prompt-responsive action control and cinematic multi-cut transitions at constant memory.
    \item We conduct extensive qualitative and quantitative evaluations on VBench, showing that \methodName achieves state-of-the-art long-video subject, background consistency,  motion smoothness and dynamic degree.
\end{itemize}

\section{Related Work}
\label{sec:related}

\begin{figure*}[t]
    \centering
        \includegraphics[width=1.0\linewidth]{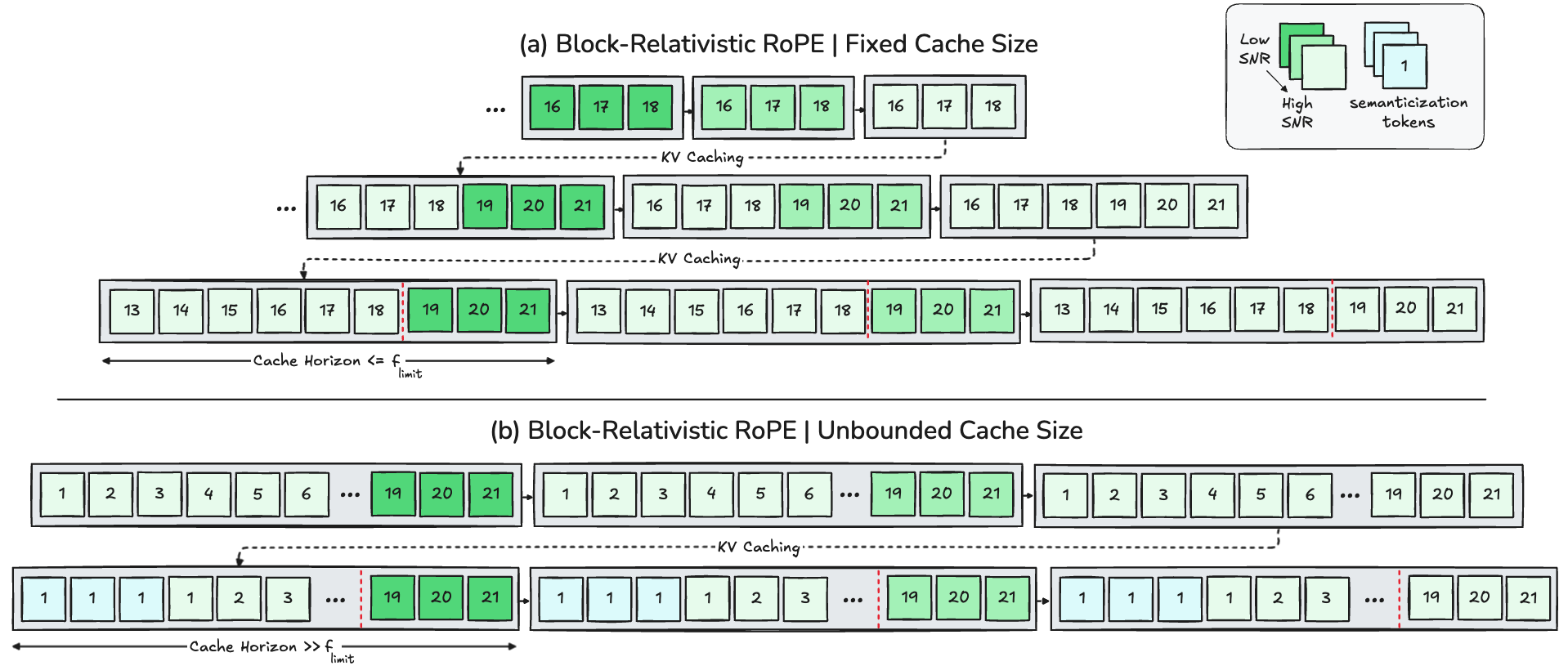} \\
    \vspace{-1em}
    \caption{\textbf{Block-Relativistic RoPE.} \textbf{(a) Fixed cache size.}
As new latent blocks are generated, their temporal RoPE coordinates are rotated relative to the teacher’s maximum horizon $f_{\text{limit}}$, while earlier latents are rotated backward to preserve their relative temporal geometry within the fixed cache window. 
\textbf{(b) Unbounded cache size.}
When the KV cache grows beyond $f_{\text{limit}}$, earlier latents undergo semanticization: Temporally distant tokens collapse into abstract semantic memory, while recent high-SNR tokens retain precise temporal geometry. See Sec.~\ref{subsec:block_relativistic_rope} for details.}
\label{fig:block_relativistic_rope}
\vspace{-1.4em}
\end{figure*}
\noindent \textbf{Bidirectional Video Generation.}
Recent years have witnessed remarkable progress in video generation, largely driven by the success of denoising diffusion models~\cite{ho2020denoising, song2020denoising, lipman2022flow}. These approaches have evolved from early pixel-space diffusion methods to more efficient latent-space formulations~\cite{ho2022imagen, singer2022make, blattmann2023align, blattmann2023stable}, supported by increasingly powerful architectures ranging from space–time U-Nets~\cite{blattmann2023stable, hong2022cogvideo} to diffusion transformers (DiT)~\cite{peebles2023scalable, gupta2024photorealistic, veo, hong2022cogvideo, yang2024cogvideox, openai_sora, wan2025wan, kong2024hunyuanvideo}. These large-scale systems typically operate as bidirectional video diffusion models, meaning they leverage information from both past and future frames during the denoising process. This bidirectional conditioning yields coherent and visually consistent results. However, their reliance on full temporal context prevents deployment in real-time scenarios, where future frames are unavailable at inference time.  

\noindent \textbf{Autoregressive Video Generation.}
SkyReels-V2~\cite{chen2025skyreels} integrates Diffusion Forcing~\cite{chen2024diffusion} with RL and a non-decreasing noise schedule for infinite-length synthesis. CausVid~\cite{yin2025slow}  distilled bidirectional DiT into causal generators using asymmetric DMD~\cite{yin2024improved, yin2024one}, achieving streaming video synthesis but limited consistency beyond short horizons. Self-Forcing~\cite{huang2025self} mitigated this train–test mismatch by conditioning on self-generated frames during training, improving temporal stability but still degrading beyond the teacher’s horizon. Self-Forcing++~\cite{cui2025self} extended this with long-horizon distillation on self-rollouts via a rolling KV cache, enabling minute-scale synthesis. Rolling Forcing~\cite{liu2025rolling} jointly denoises consecutive frames with progressively increasing noise and maintains a persistent attention-sink cache for global consistency. LongLive~\cite{yang2025longlive} complements this with frame-level autoregression and \textit{KV re-cache} tuning for  interactive generation. Stable Video Infinity (SVI)~\cite{li2025stable} bridges the train-test gap via error-recycling fine-tuning, which injects self-generated errors during training. BAgger~\cite{po2025bagger} constructs corrective trajectories by reversing the model's own rollouts. Reward Forcing~\cite{lu2025reward} enhances motion dynamics through an EMA sink mechanism and Rewarded Distribution Matching Distillation (Re-DMD), which biases the generator toward high-reward motion regions. Block Cascading~\cite{bandyopadhyay2025block} introduces a training-free acceleration strategy that parallelizes block-causal inference by allowing future blocks to begin generation conditioned on the partially denoised latents of preceding blocks. NOVA~\cite{deng2024autoregressive} performs non-quantized frame-by-frame and set-by-set prediction, removing vector quantization for continuous latent autoregression but  limited in long-context scalability.  MAGI-1~\cite{teng2025magi} adopts chunk-wise autoregressive denoising with block-causal attention and parallel chunk generation at high infrastructure cost.  
In contrast to these methods, which primarily extend temporal horizons via new training procedures, distillation strategies, or large-scale infrastructure, \methodName explores what already distilled models can achieve by reparameterizing temporal RoPE and KV caching at inference time, and can be applied in a plug-and-play fashion on top of existing Self-Forcing variants to enable effectively infinite-horizon, controllable video generation. FLEX \cite{li2026train} subsequently introduced frequency-aware RoPE modulation and antiphase noise sampling to mitigate the spectral bias in long video generation.

\section{Background}
\label{sec:background}

\subsection{Base Model}
\label{subsec:preliminaries}
Our work builds upon the pretrained Wan2.1-T2V-1.3B ~\cite{wan2025wan} distilled from Wan2.1-T2V-14B via Self-Forcing.  The model operates in a latent space encoded by a 3D Variational Auto-Encoder (VAE)~\cite{kingma2013auto}, which compresses the input video by $4\times$ along the temporal dimension and $8\times$ along each spatial dimension. Formally, each input video $\mathbf{V} \in \mathbb{R}^{\text{F} \times \text{H} \times \text{W} \times 3}$ is encoded into a latent tensor $\mathbf{x}_0$ of size $1+\left\lceil\frac{\text{F-1}}{4}\right\rceil \times \text{C} \times \frac{\text{H}}{8} \times \frac{\text{W}}{8}$, where $\text{C}$ denotes the latent channel dimension.  The model follows the Rectified Flow formulation~\cite{esser2024scaling}, where the forward process interpolates between a clean latent $\mathbf{x}_0$ and Gaussian noise $\boldsymbol{\epsilon} \sim \mathcal{N}(0, \mathbf{I})$ as $\mathbf{x}_t = (1 - t)\mathbf{x}_0 + t\boldsymbol{\epsilon}$, where $t \in [0, 1]$, and the reverse process is parameterized by a neural velocity field $v_{\theta}$ as an ordinary differential equation (ODE): $d\mathbf{x}_t = v_{\theta}(\mathbf{x}_t, t)\,dt$, where $\theta$ denotes model parameters. During inference, Euler discretization is applied over $t$ to iteratively solve the ODE and recover $\mathbf{x}_0$ from $\mathbf{x}_1$.

\subsection{3D Rotary Position Embedding (3D-RoPE)}
\label{subsec:rope}

Wan~\cite{wan2025wan} employs 3D-RoPE to encode the temporal and spatial coordinates of query and key tokens before self-attention. Let the latent features be
\[
\mathbf{x} \in \mathbb{R}^{\text{B} \times \text{S} \times \text{C}}, \quad \text{S} = \text{F}\times\text{H}\times\text{W}.
\]
For simplicity, we omit the timestep index $t$. The channel dimension is divided into three groups for the temporal, height, and width axes, such that
\[
\mathbf{x} = \big[\,\mathbf{x}_f \,\|\, \mathbf{x}_h \,\|\, \mathbf{x}_w\,\big],
\quad \frac{\text{C}}{2} = \text{C}_{\!f} + \text{C}_{\!h} + \text{C}_{\!w}.
\]
For each token at coordinates \((f,h,w)\), 3D-RoPE applies rotary embeddings along the corresponding axes and concatenates the results:
\begin{align*}
\text{RoPE}_{\text{3D}}\!\big[\mathbf{x}_{(f,h,w)}\big] 
&= 
\big[\,\text{RoPE}(\mathbf{x}_f) \,\|\, \text{RoPE}(\mathbf{x}_h) \,\|\, \text{RoPE}(\mathbf{x}_w)\,\big] \\
&= 
\big[\,\text{RoPE}(\mathbf{x}_f) \,\|\, \text{RoPE}_{\text{2D}}[\mathbf{x}_{(h,w)}]\,\big]
\end{align*}

\noindent \textbf{Temporal Extrapolation Limit.}
In Wan, each RoPE dimension is configured with a fixed maximum sequence length of 1024, which defines the representable positional range for all three axes. However, the base model is trained only on short temporal horizons, so it never adapts to RoPE indices beyond those seen during training. As a result, even when generation exceeds the model's short training horizon, rather than the full 1024-frame limit, the model enters unseen positional regimes and attention quality degrades even though the RoPE formulation itself remains valid.

\section{Methodology}
\label{sec:method}
\subsection{Block-Relativistic RoPE}
\label{subsec:block_relativistic_rope}

\noindent \textbf{Fixed Cache Size.}
We first consider the case where the KV cache horizon is smaller than or equal to the base model’s native generation horizon $f_{\text{limit}}$. In our experiments, the base model is trained on 5-second clips corresponding to $f_{\text{limit}} =21$ latent frames. During autoregressive self-rollout, video generation proceeds in blocks of three latent frames, i.e.,
\begin{equation}
\mathbf{B}_f = \{\,f-2,\,{f-1},\,{f}\,\}. 
\end{equation}
Once the cache becomes full, the oldest three latent frames are evicted and replaced by the newly generated block. In our design, although only fixed size latent frames are resident in KV cache at any time, the temporal RoPE indices assigned to the current block are not bounded by this cache size but instead by the base model’s full generation capacity of $f_{\text{limit}}$ latent frames as demonstrated in Fig.~\ref{fig:block_relativistic_rope}(a). Consequently, even as tokens are rolled within the cache, their temporal embeddings can continue to advance beyond the cache length, fully harnessing the global temporal continuity within the teacher’s horizon.
\begin{figure*}[t]
    \centering
        \includegraphics[width=1.0\linewidth]{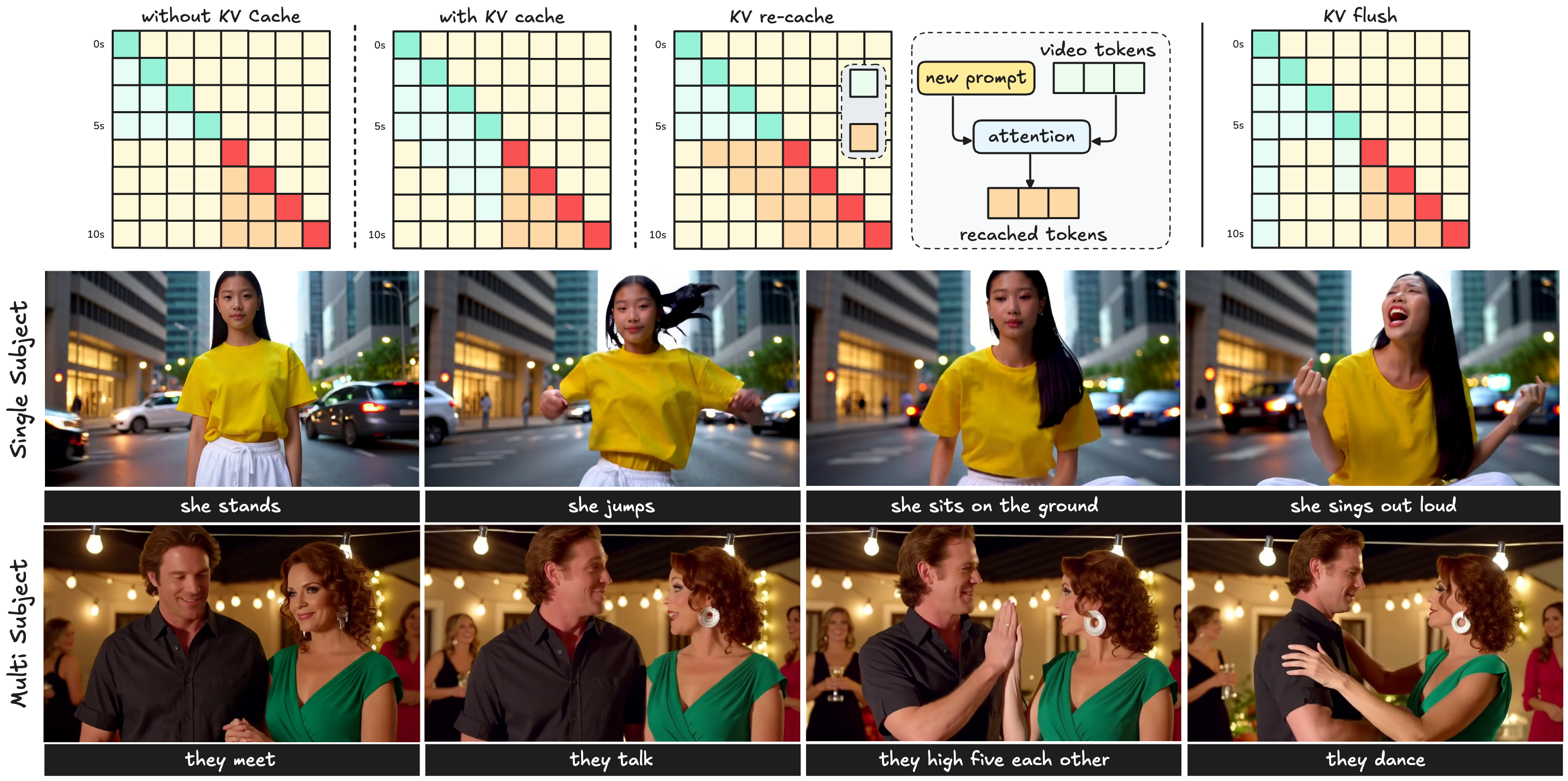} \\ \vspace{-0.8em}
    \caption{\textbf{KV Flush.}
KV Flush resets the KV cache to only two tokens, the global sink and the last latent frame, so that a new prompt takes effect immediately without carrying over old semantics. 
Compared to no-cache (abrupt changes), full-cache (semantic lag), and KV re-cache (high latency), KV Flush achieves instant, clean action responsiveness with smooth temporal continuity, as shown in the prompt sequence: standing → jumping → sitting → singing. }

\label{fig:kv_flush}
\vspace{-1.4em}
\end{figure*}
In conventional absolute $\text{RoPE}_{3\text{D}}$, temporal indices are defined globally. For temporal index $i \gg f_{\text{limit}}$ and current generation block $\mathbf{B}_i = \{i-2,\,{i-1},\,{i}\}$
\[
\text{RoPE}_{3\text{D}}\!\big[\mathbf{x}_{(\text{B}_i,h,w)}\big]
=
\big[\,\text{RoPE}(\mathbf{x}_{\text{B}_i})\,\Vert\,\text{RoPE}_{\text{2D}}[\mathbf{x}_{(h,w)}]\,\big].
\]
In contrast, Block-Relativistic RoPE defines temporal coordinates within a moving local reference frame while keeping spatial coordinates fixed:
\begin{equation}
\tilde{\mathbf{B}}_i =
\begin{cases}
\mathbf{B}_i = \{\,i-2,\, i-1,\, i\,\}, & \text{if } i \leq f_0, \\[4pt]
\mathbf{B}_{f_0} = \{\,f_0-2,\, f_0-1,\, f_0\,\}, & \text{otherwise},
\end{cases}
\end{equation}
\[
\text{RoPE}_{3\text{D}}\!\big[\mathbf{x}_{(\mathbf{B}_i, h, w)}\,\big|\,f_0\big]
=
\big[\,\text{RoPE}(\mathbf{x}_{\tilde{\mathbf{B}}_i})\,\Vert\,\text{RoPE}_{\text{2D}}[\mathbf{x}_{(h,w)}]\,\big],
\]
\noindent where $f_0 \leq f_{\text{limit}}$ denotes the onset index of the current cache window. As generation advances, previously cached latent frames are re-anchored relative to the $f_0$, effectively rotating their temporal phase backward within the RoPE space.

\noindent \textbf{Unbounded Cache Size.}  
When the KV cache extends beyond the teacher model’s temporal horizon $f_{\text{limit}}$, the temporal encoding must transition from a strictly temporal regime to a semantically abstract one. Drawing inspiration from the \textit{semanticization process}~\cite{renoult2012personal} in cognitive neuroscience, we reinterpret this transition as a form of temporal \textit{decontextualization}. In the human brain, semanticization describes how episodic memories, which are initially tied to specific times and places, gradually lose their spatiotemporal precision while preserving their conceptual and self-relevant meaning. Through consolidation, these details become integrated into the brain’s semantic networks, forming \textit{personal semantics}: knowledge of one’s past detached from its original temporal context.

Analogously, when the KV cache surpasses \(f_{\text{limit}}\), earlier latent frames undergo semanticization within the Block-Relativistic RoPE formulation as shown in Fig.~\ref{fig:block_relativistic_rope}(b). Rather than preserving unique temporal indices for the earliest frames, we collapse their temporal coordinates to a shared minimum index, effectively anchoring them as temporally invariant yet semantically influential memories. Concretely, when the current generation block is \(\mathbf{B}_i\) with \(i > f_{\text{limit}}\), the earliest cached block is remapped as
\begin{equation}
\mathbf{B}_3 = \{1,2,3\} \;\longrightarrow\; \mathbf{B}_{\bar{1}} = \{1,1,1\}.
\end{equation}
Hence, preserving \textit{what} occurred in earlier frames (their visual–semantic content) rather than \textit{when} it exactly occurred, mirroring the abstraction of episodic detail into decontextualized knowledge. 
\begin{figure*}[t]
    \centering
        \includegraphics[width=1.0\linewidth]{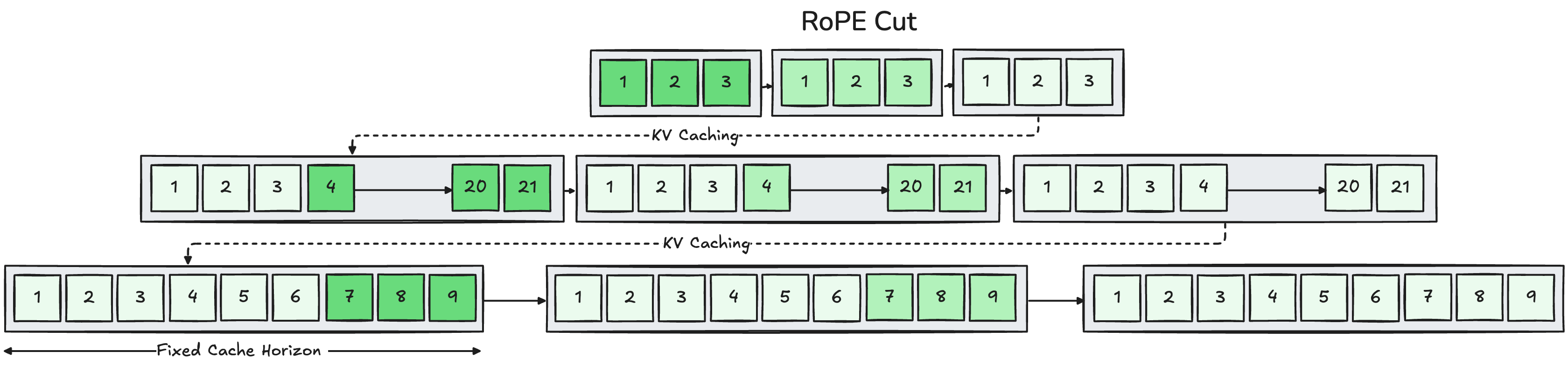} \\
    \vspace{-1em}
    \caption{\textbf{RoPE Cut.} RoPE Cut enables a discontinuous jump along the temporal RoPE axis.  In the first rollout (second row), the active latent block $B_{6}=\{4,5,6\}$ is reassigned to a new RoPE-local frame, becoming $B_{4\rightarrow 21}=\{4,20,21\}$: the token ``4'' is kept as the local anchor while the next two tokens are reassigned to the high-SNR positions $20$ and $21$ for continued denoising.  
In the subsequent rollout (third row), the block $\{4,20,21\}$ is treated as past context after the cut, and generation proceeds again from the original temporal location with a fresh block $B_{6}=\{4,5,6\}$ inside the fixed cache horizon.  
}
\label{fig:block_relativistic_rope2}
\vspace{-1.2em}
\end{figure*}
Consequently, the system’s temporal representation evolves from an \textit{episodic phase}, where frame order and timing dominate, to a \textit{semantic phase}, where distant history persists only as contextual priors shaping ongoing generation. Within Block-Relativistic RoPE, this mechanism allows the temporal axis to expand indefinitely, maintaining precise relative encoding within the current active window \(\mathbf{B}_i\) while assimilating all preceding frames into a unified, temporally agnostic semantic field.  

\begin{figure*}[t]
    \centering
    \begin{subfigure}[b]{0.26\textwidth}
        \centering
        \includegraphics[width=\linewidth]{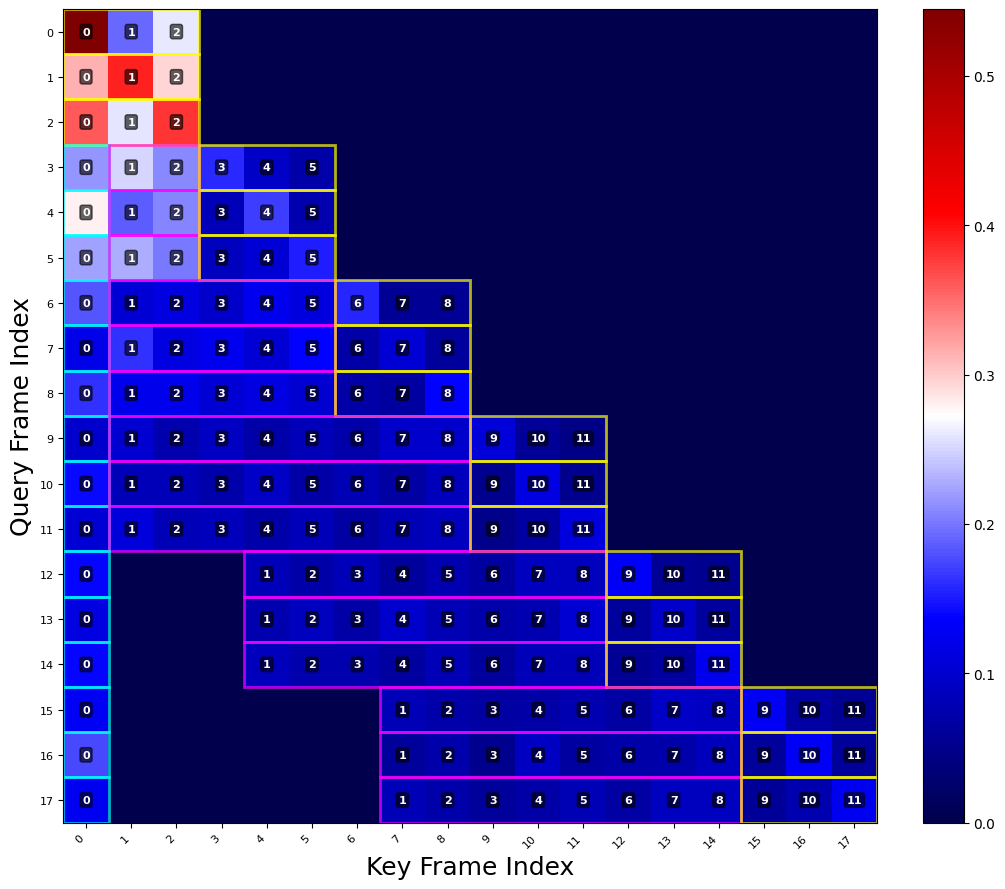}
        \caption{Block-Relativistic RoPE}
        \label{fig:attn_map_1}
    \end{subfigure}
    \hfill
    \begin{subfigure}[b]{0.34\textwidth}
        \centering
        \includegraphics[width=\linewidth]{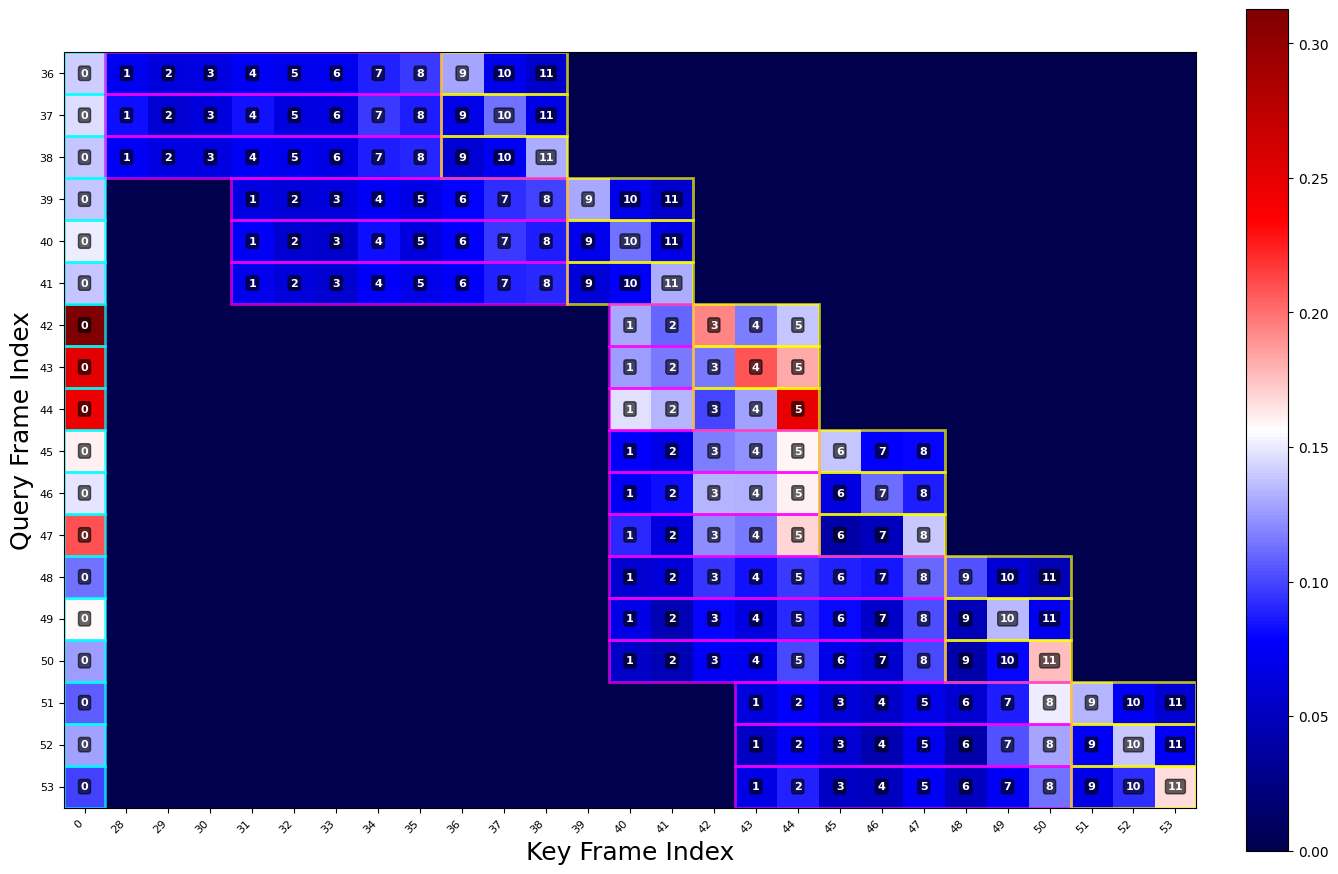}
        \caption{KV Flush}
        \label{fig:attn_map_2}
    \end{subfigure}
    \hfill
    \begin{subfigure}[b]{0.34\textwidth}
        \centering
        \includegraphics[width=\linewidth]{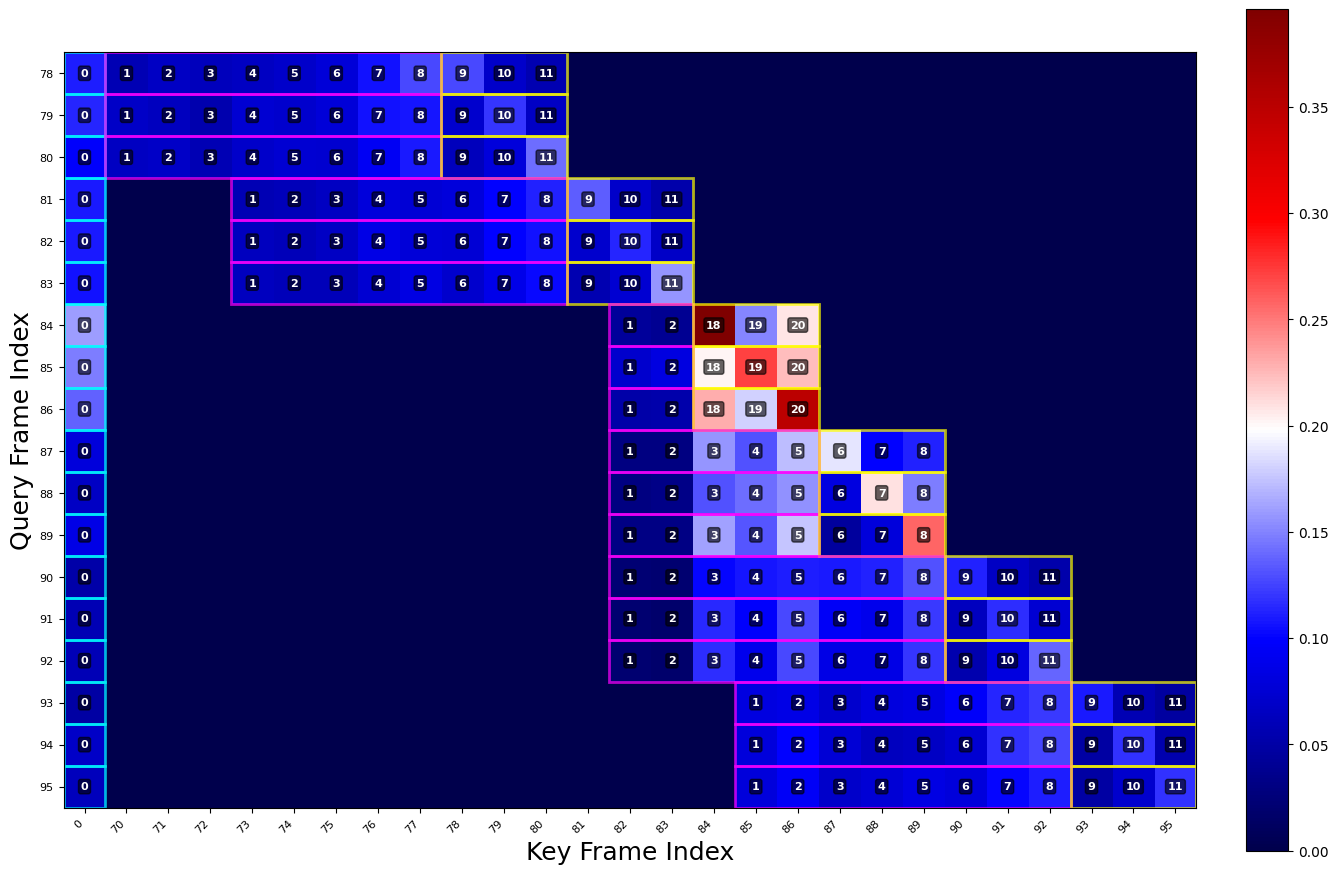}
        \caption{RoPE Cut}
        \label{fig:attn_map_3}
    \end{subfigure}
    \vspace{-1em}
    \caption{\textbf{Attention behavior of $\infty$-RoPE across interventions.} 
    Frame-to-frame attention maps from the 13th DiT layer, shown as head-averaged, 
    pixel-summed attention with query frame index $q$ on the y-axis and key frame 
    index $k$ on the x-axis, for (a) a baseline $\infty$-RoPE rollout, (b) KV Flush 
    at an action change, and (c) RoPE Cut. These visualizations summarize how our 
    method structures long-horizon temporal dependencies; see Sec.~\ref{sec:mech-attn} 
    for a detailed mechanistic interpretation.}
    \label{fig:all_attention_maps}
    \vspace{-1em}
\end{figure*}

\subsection{Action Control via KV Flush}
\label{subsec:kv_flush}
In autoregressive video diffusion, transitioning between prompts during inference, known as action-controllable video generation, requires balancing immediate semantic responsiveness with temporal continuity. \textbf{KV Flush} mechanism performs cache renewal with constant memory and zero latency as described in Fig. \ref{fig:kv_flush}. When a new prompt arrives, all cached tokens are flushed except two anchors, the global sink latent frame and the last generated latent frame, which respectively stabilize attention normalization and preserve local temporal continuity. The next action is then conditioned directly on these minimal anchors, allowing the scene to evolve smoothly in motion while instantly adopting the new semantics outperforming prior cache management paradigms in both efficiency and controllability.
\begin{table*}[t]
\centering
\caption{\textbf{Performance comparisons on 5s and 60s videos.} For 5s generations, several baselines report high temporal-quality metrics largely due to stagnation effects reflected in their low dynamic degree, whereas $\infty$-RoPE maintains strong temporal stability without sacrificing motion richness. On 60s videos, $\infty$-RoPE demonstrates superior long-horizon coherence, subject and background consistency, and overall visual quality.}
\vspace{-1em}

\label{tab:res_5s_60s}
\resizebox{\textwidth}{!}{
\begin{tabular}{lc|cccccccc|cccccccc}
\toprule
\textbf{Model} & \textbf{Throughput$\uparrow$} &
\multicolumn{8}{c}{\textbf{Results on 5s $\uparrow$}} &
\multicolumn{8}{c}{\textbf{Results on 60s $\uparrow$}} \\
\cmidrule(lr){3-10} \cmidrule(lr){11-18}
& \textbf{(FPS)} &
\shortstack{\textbf{Aesthetic} \\ \textbf{Quality}} &
\shortstack{\textbf{Background} \\ \textbf{Consistency}} &
\shortstack{\textbf{Dynamic} \\ \textbf{Degree}} &
\shortstack{\textbf{Imaging} \\ \textbf{Quality}} &
\shortstack{\textbf{Motion} \\ \textbf{Smoothness}} &
\shortstack{\textbf{Subject} \\ \textbf{Consistency}} &
\shortstack{\textbf{Temporal} \\ \textbf{Flickering}} &
\cellcolor{verylightgray}\textbf{Overall$\uparrow$} &
\shortstack{\textbf{Aesthetic} \\ \textbf{Quality}} &
\shortstack{\textbf{Background} \\ \textbf{Consistency}} &
\shortstack{\textbf{Dynamic} \\ \textbf{Degree}} &
\shortstack{\textbf{Imaging} \\ \textbf{Quality}} &
\shortstack{\textbf{Motion} \\ \textbf{Smoothness}} &
\shortstack{\textbf{Subject} \\ \textbf{Consistency}} &
\shortstack{\textbf{Temporal} \\ \textbf{Flickering}} &
\cellcolor{verylightgray}\textbf{Overall$\uparrow$} \\
\midrule

\rowcolor{lightblue}
\multicolumn{18}{l}{\textit{Bidirectional models}} \\
LTX-Video~\cite{hacohen2024ltx} & 8.98 & 0.6209 & 0.9530 & \textbf{0.64} & \underline{0.7289} & \underline{0.9941} & 0.9501 & 0.9812 & \cellcolor{verylightgray}\textbf{0.8439} & - & - & - & - & - & - & - & - \\
Wan2.1~\cite{wan2025wan} & 0.78 & \textbf{0.6675} & \underline{0.9696} & 0.44 & 0.7195 & 0.9851 & 0.9599 & 0.9745 & \cellcolor{verylightgray}0.8319 & - & - & - & - & - & - & - & - \\

\midrule

\rowcolor{lightblue}
\multicolumn{18}{l}{\textit{Autoregressive models}} \\
NOVA~\cite{deng2024autoregressive} & 0.88 & 0.5819 & 0.9516 & 0.28 & 0.6338 & 0.9908 & 0.9338 & 0.9833 & \cellcolor{verylightgray}0.7915 & 0.4753 & 0.8806 & 0.12 & 0.4497 & 0.9894 & 0.7750 & 0.9827 & \cellcolor{verylightgray}0.6901 \\
Pyramid Flow~\cite{jin2024pyramidal} & 6.70 & 0.6377 & 0.9609 & 0.44 & 0.6650 & 0.9935 & 0.9608 & \underline{0.9861} & \cellcolor{verylightgray}0.8266 & - & - & - & - & - & - & - & -\\
MAGI-1~\cite{teng2025magi} & 0.19 & 0.6360 & 0.9683 & 0.44 & 0.6022 & \textbf{0.9945} & 0.9583 & \textbf{0.9900} & \cellcolor{verylightgray}0.8199 & 0.5210 & 0.8776 & \textbf{0.56} & 0.5454 & \textbf{0.9926} & 0.7946 & \underline{0.9848} & \cellcolor{verylightgray}0.7511 \\
SkyReels-V2~\cite{chen2025skyreels} & 0.49 & 0.6621 & 0.9683 & 0.48 & 0.7016 & 0.9884 & 0.9607 & 0.9757 & \cellcolor{verylightgray}0.8335 & 0.5764 & 0.8995 & 0.44 & 0.6667 & 0.9867 & 0.8499 & 0.9760 & \cellcolor{verylightgray}0.7768 \\
CausVid~\cite{yin2025slow} & 17.01 & 0.6443 & 0.9512 & 0.48 & 0.7012 & 0.9832 & 0.9596 & 0.9730 & \cellcolor{verylightgray}0.8231 & 0.6288 & 0.8985 & \underline{0.52} & 0.6747 & 0.9847 & 0.8675 & 0.9755 & \cellcolor{verylightgray}0.7940 \\

Self-Forcing~\cite{huang2025self} & 17.01 & 0.6576 & 0.9598 & \underline{0.60} & \textbf{0.7323} & 0.9841 & 0.9629 & 0.9677 & \cellcolor{verylightgray}\underline{0.8398} & 0.5895 & 0.8784 & 0.32 & 0.6971 & 0.9890 & 0.8360 & 0.9830 & \cellcolor{verylightgray}0.7715 \\
Rolling-Forcing~\cite{liu2025rolling} & 17.01 & \underline{0.6655} & 0.9652 & 0.40 & 0.7273 & 0.9871 & \underline{0.9721} & 0.9771 & \cellcolor{verylightgray}0.8332 & \textbf{0.6350} & \underline{0.9447} & 0.36 & \textbf{0.7242} & 0.9865 & \underline{0.9409} & 0.9769 & \cellcolor{verylightgray}\underline{0.8146} \\
\textbf{\methodName (Ours)} & 17.01 & 0.6308 & \textbf{0.9720} & 0.48 & 0.7141 & 0.9872 & \textbf{0.9787} & 0.9845 & \cellcolor{verylightgray}0.8377 & \underline{0.6325} & \textbf{0.9490} & \underline{0.52} & \underline{0.7010} & \underline{0.9901} & \textbf{0.9444} & \textbf{0.9852} & \cellcolor{verylightgray}\textbf{0.8298} \\
\bottomrule
\end{tabular}
}\vspace{-1em}
\end{table*}
\subsection{Multi-Cut Scenes via RoPE Cut}
\label{subsec:rope_cut}

While Block-Relativistic RoPE and KV Flush enable continuous and controllable generation, cinematic narratives often require abrupt temporal discontinuities such as scene cuts, flashbacks, or cross-location transitions that cannot be represented by smooth temporal evolution alone. To achieve such discontinuities within a single continuous rollout, we introduce \textbf{RoPE Cut}, a training-free operation that performs controlled discontinuous jumps in temporal RoPE coordinates.

For the current generation block $\mathbf{B}_f = \{\!f\!-\!2,\,f\!-\!1,\,f\!\}$, RoPE Cut redefines the temporal coordinate mapping as
\[
\mathbf{B}_{f \rightarrow f+\Delta} = \{\!f\!-\!2,\,f\!+\!\Delta\!-\!1,\,f\!+\!\Delta\!\},
\]
where $\Delta$ denotes an offset representing the temporal gap between the current and the next scene. Instead of extending temporal indices sequentially, we cut and re-anchor the temporal RoPE phase by offsetting the current block by $\Delta$ frames:
\begin{equation}
\text{RoPE}_{3\text{D}}\!\big[\mathbf{x}_{(\mathbf{B}_f,h,w)}\big]
\;\longrightarrow\;
\text{RoPE}_{3\text{D}}\!\big[\mathbf{x}_{(\mathbf{B}_{f \rightarrow f+\Delta},h,w)}\big].
\end{equation}

As illustrated in Fig.~\ref{fig:block_relativistic_rope2} (second row), when a discontinuous jump occurs, the subsequent frames (third row) are generated as if the jumped segment were repositioned into the past temporal indices. This re-indexing allows the system to fully reuse the base model’s generation horizon $f_{\text{limit}}$ without violating its temporal range. Thanks to our relativistic formulation, there is no fixed or absolute position along the temporal axis—the coordinate system itself shifts with each cut, allowing identity preservation even after large temporal or semantic jumps.

\subsection{Mechanistic Interpretation of \methodName}
\label{sec:mech-attn}

We analyze frame-to-frame self-attention maps to interpret how \methodName structures long-horizon temporal dependencies. During long video generation via Block-Relativistic RoPE (Fig.~\ref{fig:attn_map_1}), attention forms a sharp diagonal band, showing that each frame attends primarily to its recent predecessors, and a persistent early-time \textit{sink} column that aggregates global scene information. This pattern remains stable far beyond the original 3D-RoPE horizon, indicating that the relativistic re-anchoring of RoPE preserves relative temporal geometry and prevents attention collapse. Interventions reshape this structure in a controllable way. With KV Flush (Fig.~\ref{fig:attn_map_2}), attention from new frames to intermediate past frames is largely suppressed, and mass is redirected toward the sink and the last few pre-flush frames, enabling instant prompt responsiveness while maintaining local motion continuity. With RoPE Cut (Fig.~\ref{fig:attn_map_3}), the map splits into two nearly disjoint diagonal blocks, with the new segment attending only to itself and the sink, effectively severing temporal context and realizing a clean scene cut. Overall, these patterns provide a mechanistic explanation of how \methodName supports stable long-horizon rollouts, prompt-responsive control, and cinematic transitions within a single autoregressive generation.

\begin{figure}
    \centering
    \includegraphics[width=\linewidth]{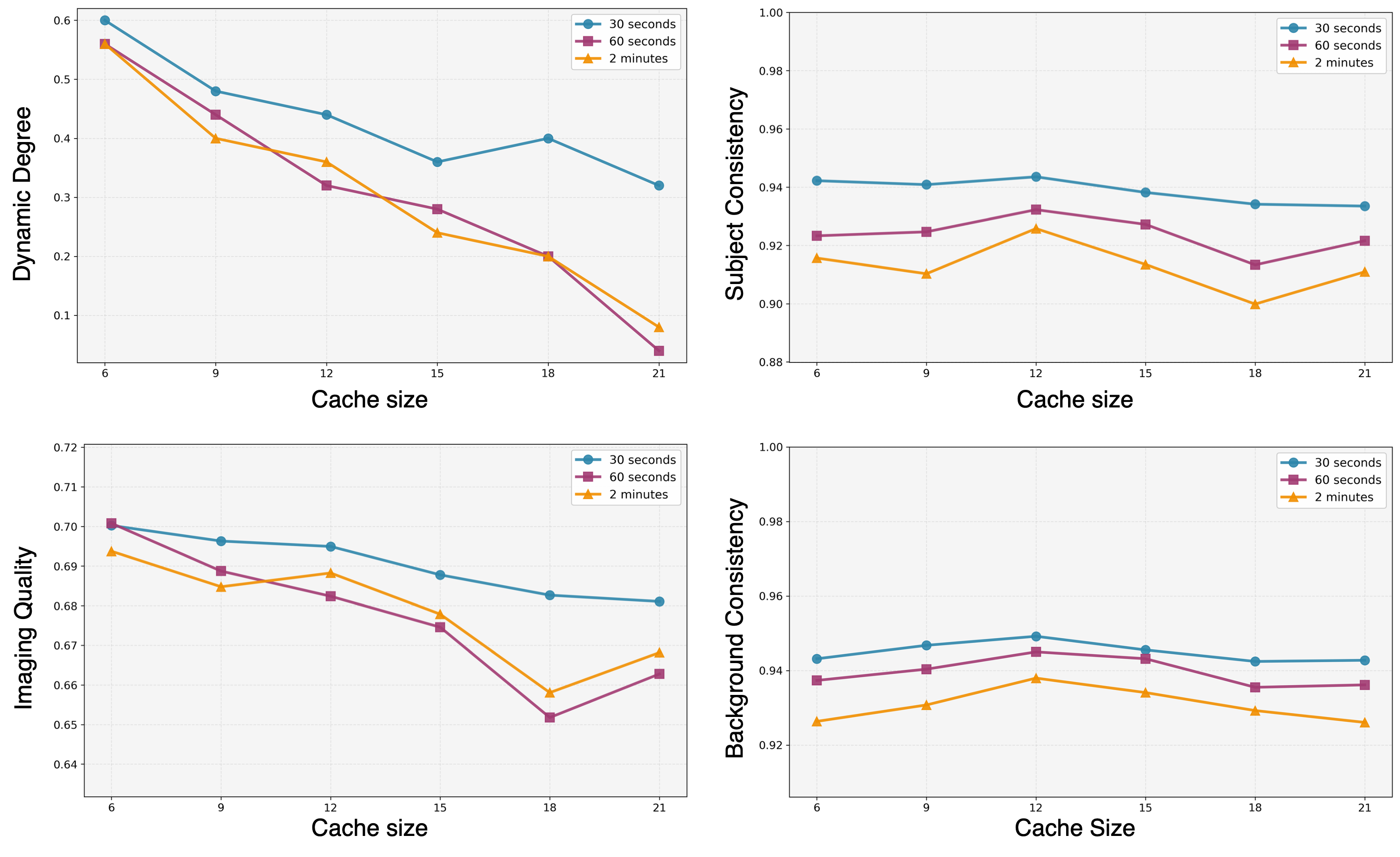}\vspace{-1em}
    \caption{\textbf{Cache Size vs. Performance across VBench dimensions.} Metrics are shown as a function of KV Cache size for (a) Overall, (b) Aesthetic Quality, (c) Dynamic Degree, and (d) Imaging Quality, evaluated on 30s, 60s and 120s sequences.}
    \label{fig:vbench_metrics}
    \vspace{-2em}
\end{figure}

\section{Experiments}
\label{sec:exp}
\begin{figure*}[t]
    \centering
    \begin{tabular}{c}
        \includegraphics[width=\linewidth]{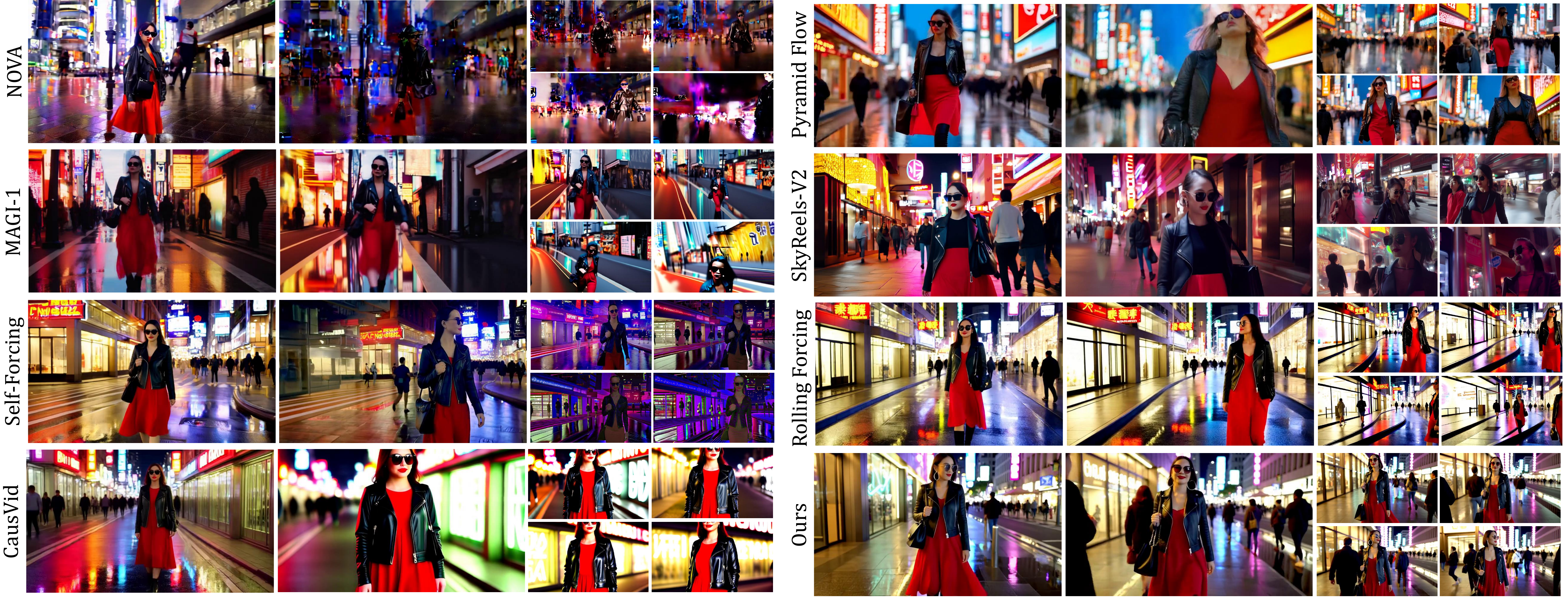} \\
    \end{tabular}
    \vspace{-1.3em}
    \caption{\textbf{Qualitative Comparison on 60-Second Generation.}
We present 60-second generations. While prior models exhibit identity drift, temporal inconsistencies, or notable degradation in visual fidelity, our approach produces sequences that remain highly subject consistent and temporally coherent, with sustained image quality across the entire one-minute horizon.}
\label{fig:qualitative}
\vspace{-1.5em}
\end{figure*}

\begin{table*}[t]
\centering
\caption{\textbf{Performance comparisons on 120s and 240s videos.} Across both horizons, $\infty$-RoPE consistently ranks first or second in every metric and achieves the strongest overall score, demonstrating stable long-horizon quality, high subject and background consistency, and robust motion dynamics over extended rollouts.}
\vspace{-1em}
\label{tab:res_120s_240s}
\resizebox{\textwidth}{!}{
\begin{tabular}{l|cccccccc|cccccccc}
\toprule
\textbf{Model} &
\multicolumn{8}{c}{\textbf{Results on 120s $\uparrow$}} &
\multicolumn{8}{c}{\textbf{Results on 240s $\uparrow$}} \\
\cmidrule(lr){2-9} \cmidrule(lr){10-17}
&
\shortstack{\textbf{Aesthetic} \\ \textbf{Quality}} &
\shortstack{\textbf{Background} \\ \textbf{Consistency}} &
\shortstack{\textbf{Dynamic} \\ \textbf{Degree}} &
\shortstack{\textbf{Imaging} \\ \textbf{Quality}} &
\shortstack{\textbf{Motion} \\ \textbf{Smoothness}} &
\shortstack{\textbf{Subject} \\ \textbf{Consistency}} &
\shortstack{\textbf{Temporal} \\ \textbf{Flickering}} &
\cellcolor{verylightgray}\textbf{Overall$\uparrow$} &
\shortstack{\textbf{Aesthetic} \\ \textbf{Quality}} &
\shortstack{\textbf{Background} \\ \textbf{Consistency}} &
\shortstack{\textbf{Dynamic} \\ \textbf{Degree}} &
\shortstack{\textbf{Imaging} \\ \textbf{Quality}} &
\shortstack{\textbf{Motion} \\ \textbf{Smoothness}} &
\shortstack{\textbf{Subject} \\ \textbf{Consistency}} &
\shortstack{\textbf{Temporal} \\ \textbf{Flickering}} &
\cellcolor{verylightgray}\textbf{Overall$\uparrow$} \\
\midrule

\rowcolor{lightblue}
\multicolumn{17}{l}{\textit{Autoregressive models}} \\
NOVA~\cite{deng2024autoregressive} & 0.4563 & 0.8629 & 0.16 & 0.4294 & \textbf{0.9908} & 0.7394 & \textbf{0.9845} & \cellcolor{verylightgray}0.6785 & 0.4345 & 0.8533 & 0.24 & 0.3854 & \underline{0.9897} & 0.7116 & 0.9827 & \cellcolor{verylightgray}0.6662 \\
SkyReels-V2~\cite{chen2025skyreels} & 0.5136 & 0.8571 & 0.40 & 0.6154 & 0.9856 & 0.7700 & 0.9765 & \cellcolor{verylightgray}0.7326 & 0.4787 & 0.8421 & \underline{0.40} & 0.6005 & 0.9841 & 0.7384 & 0.9750 & \cellcolor{verylightgray}0.7147 \\
CausVid~\cite{yin2025slow} & \underline{0.6296} & 0.8902 & 0.40 & 0.6754 & 0.9847 & 0.8485 & 0.9757 & \cellcolor{verylightgray}0.7799 & \textbf{0.6300} & 0.8837 & 0.36 & 0.6876 & 0.9845 & 0.8399 & 0.9759 & \cellcolor{verylightgray}0.7758 \\
Self-Forcing~\cite{huang2025self} & 0.5135 & 0.8321 & 0.32 & 0.6496 & 0.9877 & 0.7528 & 0.9829 & \cellcolor{verylightgray}0.7271 & 0.4286 & 0.7973 & 0.24 & 0.6141 & 0.9881 & 0.6893 & \textbf{0.9853} & \cellcolor{verylightgray}0.6850 \\
Rolling-Forcing~\cite{liu2025rolling} & \textbf{0.6319} & \underline{0.9374} & \underline{0.44} & \textbf{0.7242} & 0.9862 & \underline{0.9272} & 0.9769 & \cellcolor{verylightgray}\underline{0.8162} & 0.6156 & \underline{0.9248} & \underline{0.40} & \textbf{0.7128} & 0.9855 & \underline{0.9080} & 0.9753 & \cellcolor{verylightgray}\underline{0.8017} \\
\textbf{\methodName (Ours)} & 0.6199 & \textbf{0.9392} & \textbf{0.56} & \underline{0.6904} & \underline{0.9893} & \textbf{0.9302} & \underline{0.9835} & \cellcolor{verylightgray}\textbf{0.8236} & \underline{0.6235} & \textbf{0.9361} & \textbf{0.64} & \underline{0.7028} & \textbf{0.9898} & \textbf{0.9256} & \underline{0.9831} & \cellcolor{verylightgray}\textbf{0.8309} \\
\bottomrule
\end{tabular}
}
\vspace{-1.3em}
\end{table*}

\noindent\textbf{Implementation Details.} We implement \methodName on Self-Forcing~\cite{huang2025self}, a causal four-step generator distilled from Wan2.1-T2V-1.3B~\cite{wan2025wan} that natively produces five-second videos at 16 FPS with a resolution of \(832 \times 480\). All quantitative experiments use a KV cache size of 6, an onset index of \(f_0 = 21\), a classifier-free guidance scale of 3.0, and a timestep shift of 5.0.

\noindent \textbf{Baselines.} We compare our method against state-of-the-art open-source bidirectional and autoregressive video generation models. Specifically, we include 2 bidirectional diffusion models, Wan2.1-1.3B \cite{wan2025wan} and LTX-Video \cite{hacohen2024ltx}, and 7  autoregressive models, including Pyramid Flow \cite{jin2024pyramidal}, NOVA \cite{deng2024autoregressive}, SkyReels-V2 \cite{chen2025skyreels}, MAGI-1 \cite{teng2025magi}, CausVid \cite{yin2025slow}, Self-Forcing \cite{huang2025self}, and Rolling-Forcing \cite{liu2025rolling}.

\noindent\textbf{Evaluation.} Following~\cite{liu2025rolling, cui2025self, huang2025self}, we evaluate \methodName using VBench~\cite{huang2023vbench}, which measures subject consistency, background consistency, motion smoothness, temporal flickering, dynamic degree, aesthetic quality, and imaging quality. We randomly sample prompts from MovieGenBench~\cite{polyak2024movie} and generate over 100 videos across four durations: 5, 60, 120, and 240 seconds.

\subsection{Qualitative Experiments} 
\noindent \textbf{Qualitative Results on Long Video Generation.} We provide qualitative comparisons with previous state-of-the-art approaches in \cref{fig:qualitative}. Visually, \methodName produces frames that are notably sharper and more temporally coherent than baseline methods. A key advantage of our model is its ability to maintain superior identity consistency throughout the entire video duration, successfully mitigating the temporal drift and identity loss often observed in extended autoregressive rollouts. Furthermore, our method excels at generating highly dynamic scenes, a key advantage compared to other methods. This is quantitatively supported by our state-of-the-art Dynamic Degree scores, particularly in the long-horizon 60s 120s and 240s benchmarks (\cref{tab:res_120s_240s}). This sustained quality can be observed in both continuous long-form generation and in complex multi-prompt scenarios.

\noindent \textbf{Qualitative Results on Fine-Grained Action Control.}
Figure~\ref{fig:teaser} and Figure~\ref{fig:kv_flush} show that \methodName enables precise action control in single-subject and multi-subject streaming. Given a sequence of evolving action prompts such as \emph{standing} $\rightarrow$ \emph{jumping} $\rightarrow$ \emph{sitting} $\rightarrow$ \emph{singing}, the model produces continuous rollouts in which the subjects respond immediately to each new instruction while maintaining appearance, pose style, and scene layout across long horizons, for both single subject and multi-subject scenes. KV Flush plays a central role by resetting the cache to only the global sink token and the most recent latent frame, which removes stale context from earlier segments and allows the new action to take effect instantly.

\noindent \textbf{Qualitative Results on Cinematic Transitions.} For cinematic scene transitions, we apply RoPE Cut to introduce controlled discontinuities in the temporal RoPE coordinates while keeping the autoregressive rollout continuous shown in Fig.~\ref{fig:teaser} and Fig.~\ref{fig:appendix_scene_cut}. Qualitatively, this enables multi-cut compositions such as indoor-to-outdoor changes, time-of-day shifts, or cross-location jumps within a single generated video: the background, lighting, and context change abruptly at the cut, yet the main character’s identity, clothing, and coarse pose remain coherent before and after the transition, even when new actors enter the scene. 

\subsection{Quantitative Experiments} 
 
\noindent \textbf{Quantitative Results on Long Video Generation.}  We evaluate \methodName on both short-horizon (5 seconds) and long-horizon (60, 120, and 240 seconds) video generation using the MovieGenBench~\cite{polyak2024movie} prompts. As shown in Tab.~\ref{tab:res_5s_60s}-\ref{tab:res_120s_240s}, our method achieves state-of-the-art results compared to baselines, particularly in subject consistency and motion smoothness. Since \methodName is not limited by the RoPE dimension~\cite{cui2025self}, it maintains stable identity representation across arbitrarily long sequences. This architectural advantage explains its sustained quality in extended generations such as the 120- and 240-second benchmarks. 
\begin{figure*}[t]
    \centering
    \begin{tabular}{c}
        \includegraphics[width=\linewidth]{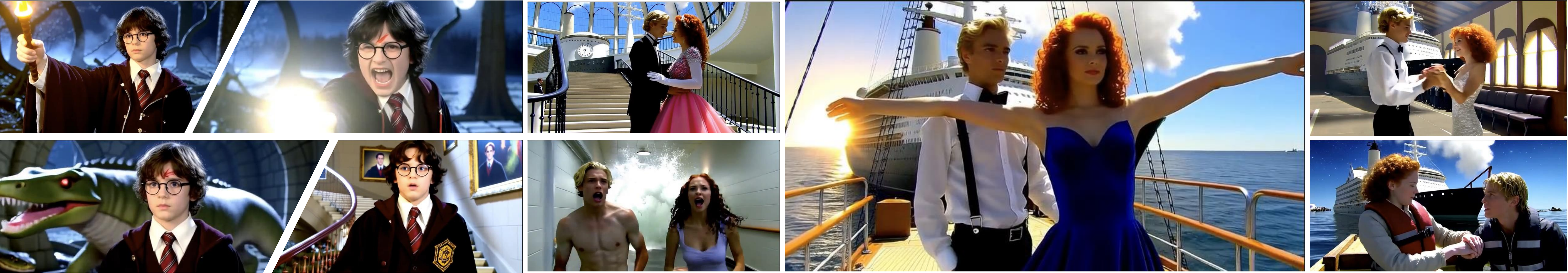} \\
    \end{tabular}
    \caption{\textbf{Dynamic Scene Cut.} RoPE Cut enables controlled cinematic transitions by introducing discontinuities in temporal RoPE coordinates. This allows a single autoregressive rollout to produce diverse environments and background changes while preserving subject identity and temporal coherence.}
\label{fig:appendix_scene_cut}
\vspace{-1.5em}
\end{figure*}
\begin{figure}
    \centering
    \includegraphics[width=\linewidth]{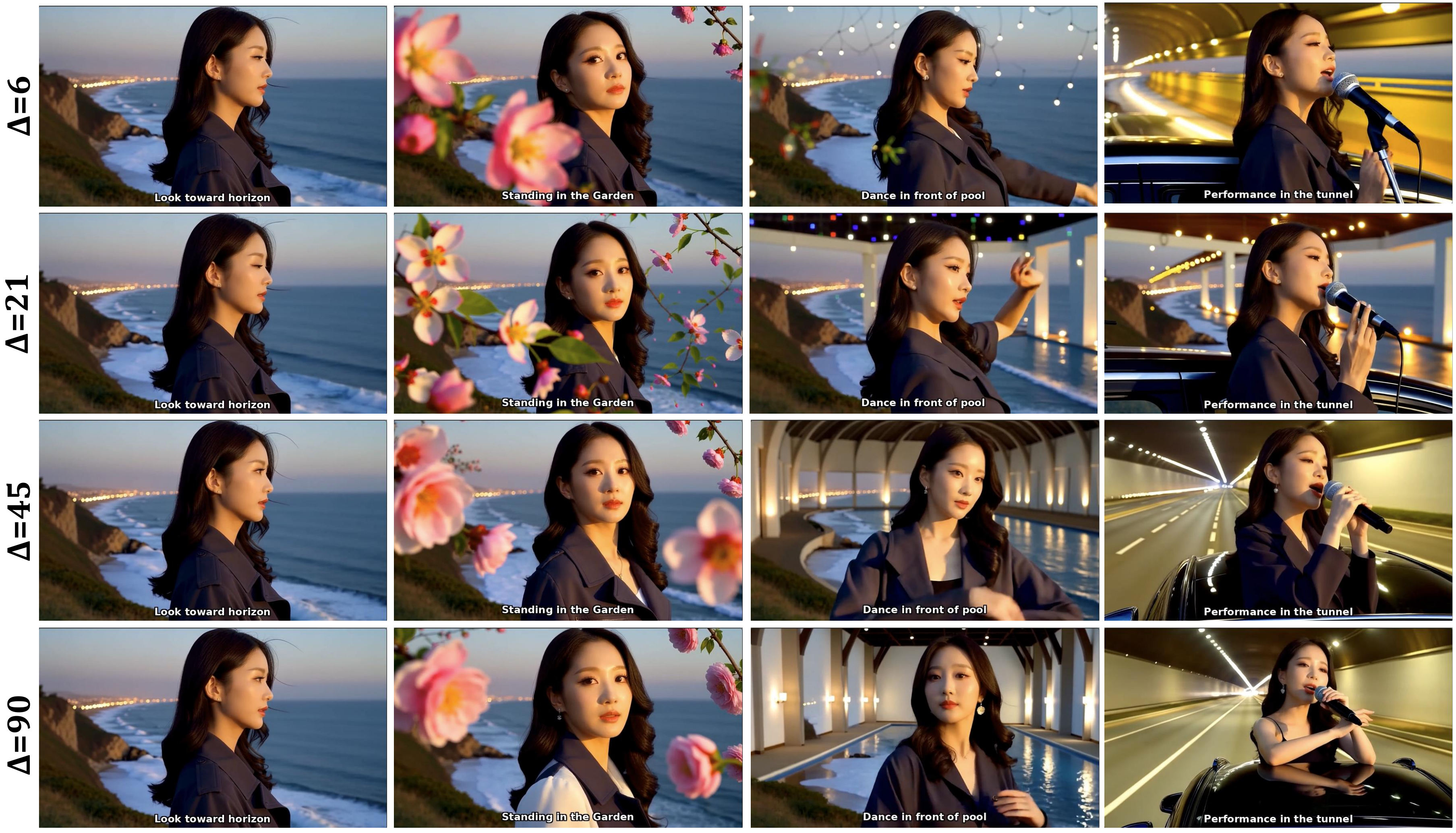}
    \caption{\textbf{Temporal Jump Index $\Delta$ Ablation.}}
    \label{fig:temporal_jump_index}
\end{figure}

\noindent \textbf{Quantitative Results on Action Controlled Video Generation.} Beyond autoregressive baselines, we also conduct a user study to compare \methodName with LongLive~\cite{yang2025longlive}, SkyReels-V2~\cite{chen2025skyreels}, and Self-Forcing~\cite{huang2025self} for action-controlled video generation in Table~\ref{tab:user_study_action}. LongLive introduces \emph{KV-Recache}, a cache management mechanism designed to enable prompt-dependent action transitions in autoregressive models. At each transition point, KV-Recache extracts all cached tokens, applies cross-attention with the new prompt, rebuilds a modified cache, and conditions subsequent frames on these recached latent tokens as demonstrated in Fig.~\ref{fig:kv_flush}. However, the approach presents two main limitations: (1) During long rollouts, embeddings from earlier prompts accumulate in the cache and are not fully erased. This reduces prompt responsiveness over time and increases the delay before the new action appears. The qualitative comparison can be found in the project page. (2) KV-Recache requires reconstructing the entire cache at every prompt change. The resulting overhead increases with the cache size and the number of action transitions. By contrast, \methodName achieves instant prompt responsiveness by simply flushing the stale cache content, which is implemented as a local update to the cache end index without proportional computation. As shown in Table ~\ref{tab:user_study_action}, \methodName yields the best Text Alignment, Subject Consistency, Motion Smoothness and Video Quality scores.
\begin{table}[t]
\centering
\scriptsize
\begin{tabular}{lcccc}
\toprule
\shortstack{\textbf{Method}} &
\shortstack{\textbf{Text}\\\textbf{Align.} $\uparrow$} &
\shortstack{\textbf{Subject}\\\textbf{Consist.} $\uparrow$} &
\shortstack{\textbf{Motion}\\\textbf{Smoothness} $\uparrow$} &
\shortstack{\textbf{Video}\\\textbf{Quality} $\uparrow$} \\
\midrule
\rowcolor{lightblue}
\multicolumn{5}{l}{\textit{Autoregressive models}} \\
Self Forcing   & 1.88 & 2.00 & 1.81 & 1.64 \\
SkyReels-V2    & 2.21 & 2.12 & 2.14 & 1.81 \\
LongLive       & 3.19 & 3.29 & 3.10 & 2.98 \\
\rowcolor{verylightgray}
\textbf{Ours}  & \textbf{3.86} & \textbf{3.95} & \textbf{3.74} & \textbf{3.38} \\
\bottomrule
\end{tabular}
\caption{\textbf{User Study on Action Controlled Video Generation.} \methodName consistently obtain higher Text Alignment, Subject Consistency, Motion Smoothness and Video Quality scores.}
\vspace{-2em}
\label{tab:user_study_action}
\end{table}

\noindent \textbf{Ablation on KV Cache Size.}
We analyze the impact of KV cache size on Dynamic Degree, Imaging Quality, Subject Consistency, and Background Consistency in Fig.~\ref{fig:vbench_metrics}. In all experiments, we fix $f_{0}=21$ and vary only the cache size. The results show that while Imaging Quality and Dynamic Degree gradually decrease as the cache grows, both Subject Consistency and Background Consistency remain stable, indicating that \methodName effectively preserves long-range identity and scene structure. 

\noindent \textbf{Ablation on Temporal Jump Index $\Delta$.} We evaluate $\Delta \in \{6, 21, 45, 90\}$. Notably, $\Delta = 6$ and $\Delta = 21$ lie within the training horizon of the pretrained model, whereas $\Delta = 45$ and $\Delta = 90$ fall outside that range. For in-horizon values, transitions remain smooth with minimal artifacts. For out-of-horizon values, the model produces more dramatic scene transitions at the cost of a visible transition edge. This edge effect arises because the block $\mathbf{B}_{f \rightarrow f + \Delta}$ is rotated by a RoPE angle that does not appear in the training data and must therefore be extrapolated. The resulting artifact reflects the inherent extrapolation behavior of the base model’s RoPE formulation. In Table~\ref{tab:delta_ablation}, we present the quantitative results for temporal jump indices $\Delta \in \{6, 21, 45, 90\}$ and qualitative results in Fig.~\ref{fig:temporal_jump_index} and Fig.~\ref{fig:appendix_scene_cut}. The results show that increasing the temporal jump index leads to lower background consistency due to the more abrupt scene transitions. However, this reduction occurs while subject consistency and temporal smoothness remain high, indicating that the model preserves identity and motion stability even under large scene changes.

\begin{table}[t]
\centering
\scriptsize
\begin{tabular}{lccc}
\toprule
\shortstack{\textbf{Temporal Jump}\\\textbf{Index $\Delta$}} &
\shortstack{\textbf{Subject}\\\textbf{Consistency} $\uparrow$} &
\shortstack{\textbf{Background}\\\textbf{Consistency} $\uparrow$} &
\shortstack{\textbf{Temporal}\\\textbf{Smoothness} $\uparrow$} \\
\midrule
$\Delta = 6$  & 90.74 & 88.98 & 0.98 \\
$\Delta = 21$ & 90.04 & 87.57 & 0.98 \\
$\Delta = 45$ & 88.47 & 84.48 & 0.96 \\
$\Delta = 90$ & 87.69 & 82.27 & 0.95 \\
\bottomrule
\end{tabular}
\caption{\textbf{Ablation Study on Temporal Jump Index $\Delta$.} We evaluate Subject Consistency, Background Consistency, and Temporal Smoothness for different values of $\Delta$ on 40-second video generation. A total of 20 videos are generated, and each video undergoes a scene cut every 10 seconds. Smaller jump values ($\Delta = 6$ and $21$) fall within the training horizon and yield smoother transitions, while larger jump values ($\Delta = 45$ and $90$) produce more pronounced scene changes accompanied by stronger transition-edge artifacts.}
\label{tab:delta_ablation}
\vspace{-2em}
\end{table}

\begin{table}[t]
\centering
\label{tab:user_study}
\begin{tabular}{l|ccc}
\toprule
\textbf{Model} &
\shortstack{\textbf{Overall} \\ \textbf{Quality}} &
\shortstack{\textbf{Temporal} \\ \textbf{Consistency}} &
\cellcolor{verylightgray}\textbf{Avg.$\uparrow$} \\
\midrule
\rowcolor{lightblue}
\multicolumn{4}{l}{\textit{Autoregressive models}} \\
CausVid~\cite{yin2025slow}         & 3.113 & 3.131 & \cellcolor{verylightgray}3.122 \\
NOVA~\cite{deng2024autoregressive} & 1.333 & 1.286 & \cellcolor{verylightgray}1.310 \\
SkyReels-V2~\cite{chen2025skyreels} & 2.579 & 2.175 & \cellcolor{verylightgray}2.377 \\
Self-Forcing~\cite{huang2025self}  & 2.458 & 2.087 & \cellcolor{verylightgray}2.273 \\
Rolling-Forcing~\cite{liu2025rolling} & \underline{3.554} & \underline{3.423} & \cellcolor{verylightgray}\underline{3.488} \\
\textbf{\methodName (Ours)}        & \textbf{3.911} & \textbf{3.708} & \cellcolor{verylightgray}\textbf{3.810} \\
\bottomrule
\end{tabular}
\caption{\textbf{User Study.} Evaluating 60 second video generations on a 5 point Likert scale, with higher scores reflecting better performance.}
\vspace{-1em}
\end{table}

\noindent \textbf{Long Video Generation User Study.} We conducted a user study with 50 participants recruited from Prolific, who rated each video independently on a 5-point Likert scale for two aspects: (i) overall quality given the prompt, and (ii) temporal consistency from start to end. We report the mean scores across all prompts and videos for each model. \methodName achieves the highest average overall quality (3.91) and temporal consistency (3.71), outperforming the strongest baseline, Rolling-Forcing (3.55 / 3.42), as well as CausVid, SkyReels-V2, Self-Forcing, and NOVA. These results align with our automatic metrics and indicate that users perceive \methodName’s long videos as both higher quality and more temporally stable.

\section{Limitations and Conclusion}

As a training-free method, \methodName inherits limitations of its base model, such as imperfect physics. Nonetheless, our results show that a simple training-free mechanism can significantly extend autoregressive video diffusion models without additional data, model updates. This work takes a practical step toward user-controllable long video generation and future scalable, temporally robust video models.

\newpage

{
    \small
    \bibliographystyle{ieeenat_fullname}
    \bibliography{main}
}

\newpage
\appendix
\maketitlesupplementary

\section{Videos and Website}
\label{sec:website}
To support thorough evaluation and improve the accessibility of our findings, we provide more than one hundred video results totaling over two hours of content. These include motivation examples, detailed qualitative demonstrations, ablation studies, and side by side comparisons. All videos are available on our Project Page: \normalsize{\url{https://infinity-rope.github.io}}.

\begin{figure}
    \centering
    \includegraphics[width=\linewidth]{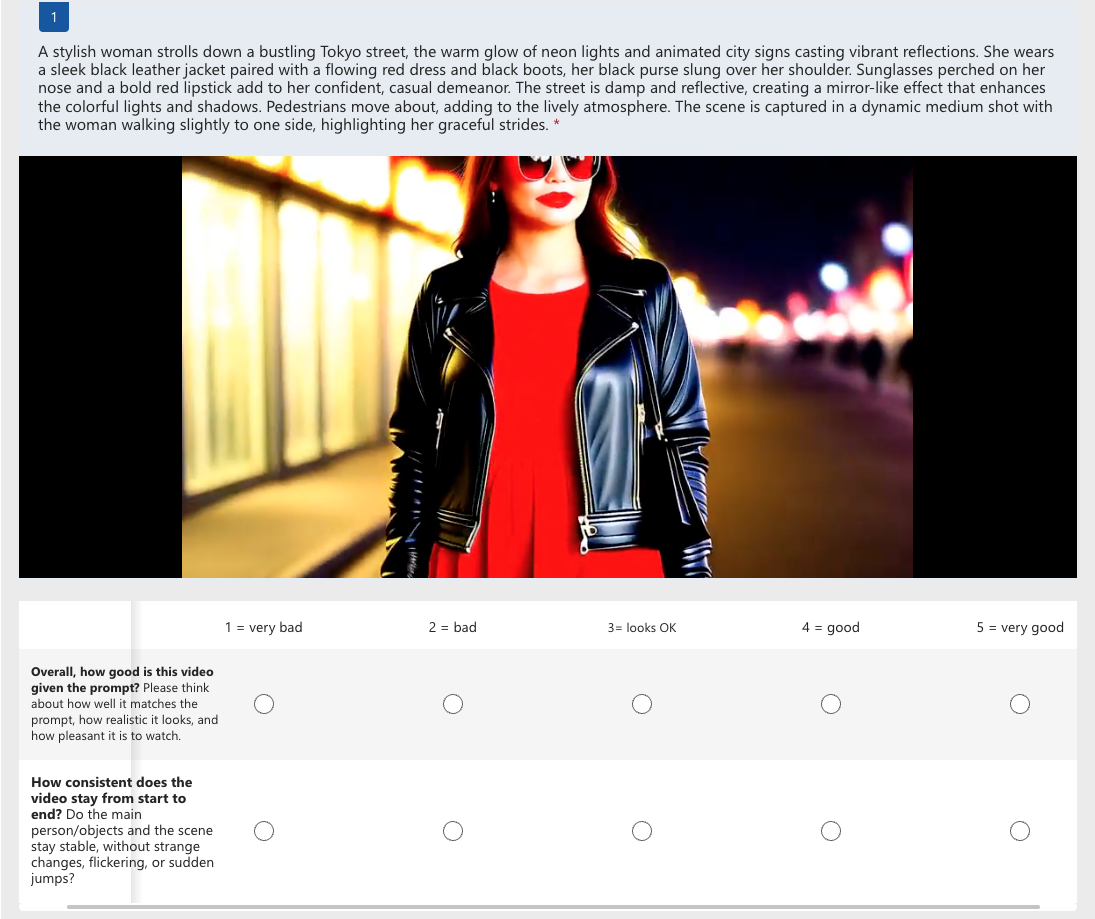}
    \caption{\textbf{User Study Interface.} User Study Interface for Long Video Generation}
    \label{fig:user_study_long}
\end{figure}

\begin{figure}
    \centering
    \includegraphics[width=\linewidth]{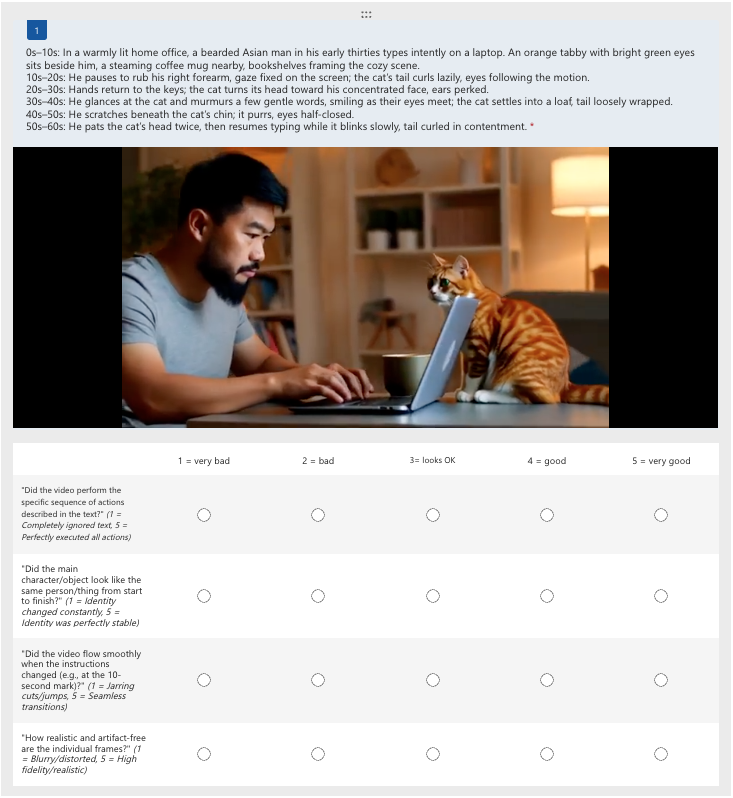}
    \caption{\textbf{User Study Interface.} User Study Interface for Action Controlled Long Video Generation}
    \label{fig:user_study_interactive}
    \vspace{-2em}
\end{figure}

\begin{figure*}[t]
    \centering
    \begin{tabular}{c}
        \includegraphics[width=\linewidth]{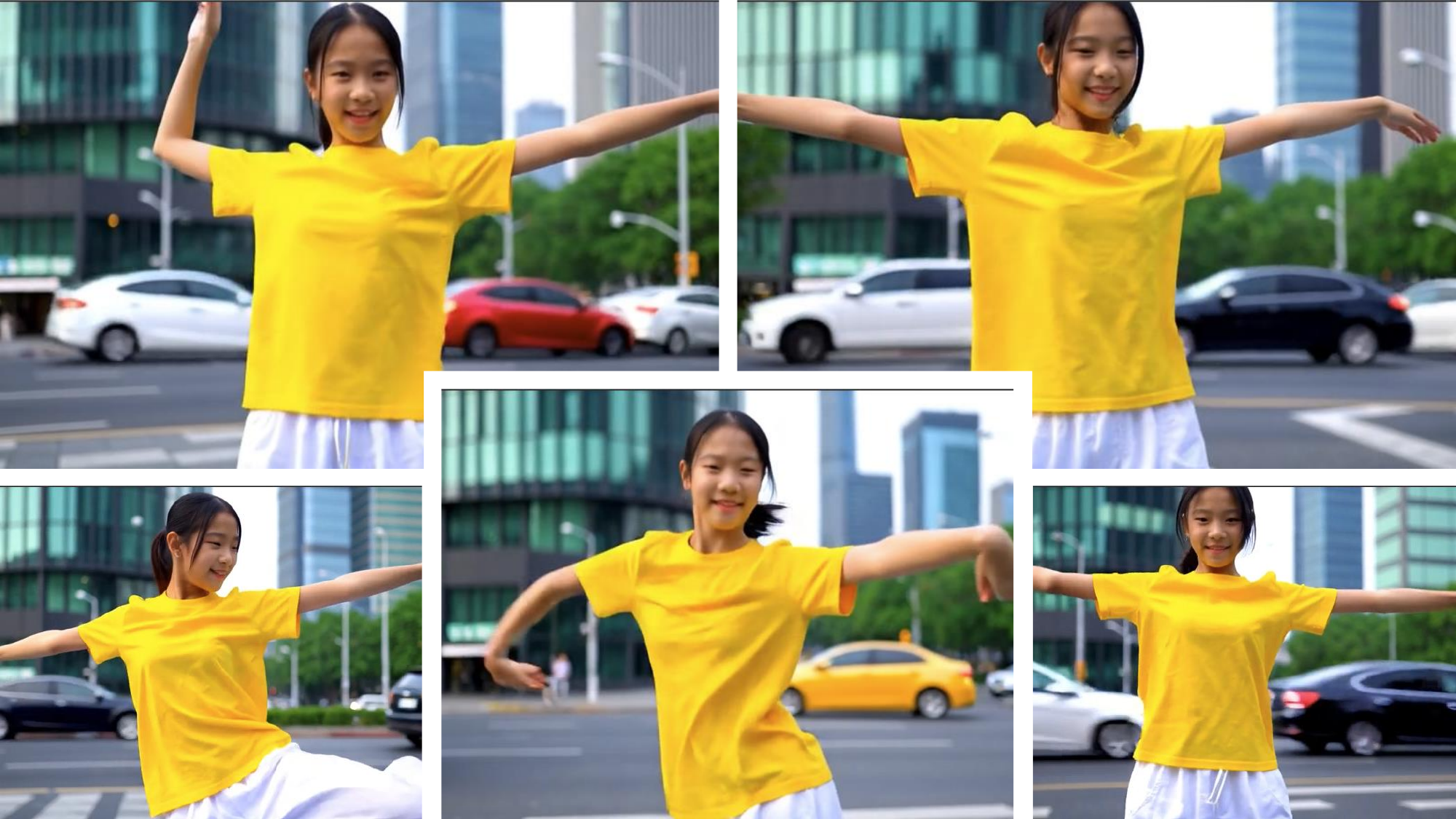} \\
    \end{tabular}
    \caption{\textbf{Ultra-Long Video Generation Enabled by Block-Relativistic RoPE.} Block-Relativistic RoPE reformulates temporal encoding as a moving frame of reference, allowing the model to preserve relative temporal geometry far beyond the base model’s positional horizon. This enables continuous, stable, and fully coherent video generation over extremely long durations without retraining or increased cache size.}

\label{fig:appendix_ultra_long}
\vspace{-1.5em}
\end{figure*}

\section{User Studies}
In Figure~\ref{fig:user_study_long} and Figure~\ref{fig:user_study_interactive}, we present the interfaces used in our two user studies. For the long-form video generation user study, participants were shown videos generated from provided prompts and were asked two questions: "Overall, how good is this video given the prompt?" and "How consistent does the video stay from start to end?" These questions were designed to evaluate prompt adherence and temporal consistency. For the action-controlled video generation user study, we compare our method against LongLive~\cite{yang2025longlive}, SkyReels-V2~\cite{chen2025skyreels}, and Self-Forcing~\cite{huang2025self}. In this study, participants were asked four questions: "Did the video perform the specific sequence of actions described in the text?" to measure prompt responsiveness, "Did the main character or object look like the same person or thing from start to finish?" to assess subject consistency, "Did the video flow smoothly when the instructions changed?" to examine motion smoothness at action transition points, and "How realistic and artifact-free are the individual frames?" to evaluate overall video quality.

\section{More Discussion on Qualitative Results}
\label{sec:qualitative_discussion}

\subsection{Discussion on Long Video Generation Results}
\label{sec:action_control_comparison}

In the main paper, we compare \methodName against NOVA~\cite{deng2024autoregressive}, MAGI-1~\cite{teng2025magi}, SkyReels-V2~\cite{chen2025skyreels}, CausVid~\cite{yin2025slow}, Self-Forcing~\cite{huang2025self}, and Rolling-Forcing~\cite{liu2025rolling} across both short (5\,s) and long (60\,s, 120\,s, 240\,s) generation settings. Our quantitative evaluations consistently show that, in the long-duration regime, \methodName outperforms prior autoregressive approaches in terms of \text{Subject Consistency}, \text{Background Consistency}, and \text{Dynamic Degree}, while ranking first or second in \text{Motion Smoothness} and \text{Temporal Flickering}.

To validate that these quantitative trends align with perceptual quality, we provide \textbf{project page} with qualitative comparisons at all four durations. The qualitative results corroborate the numerical findings. As rollout length increases, Rolling-Forcing tends to repeatedly regenerate/spawn similar characters with minimal scene evolution, a limitation stemming from its training paradigm. SkyReels-V2 exhibits large, unstable camera motions that reduce subject consistency in long sequences. Pyramidal Flow frequently resets scene content every 5 seconds, resulting in low subject and background continuity. Meanwhile, both CausVid and Self-Forcing gradually accumulate exposure bias in extended rollouts.

In contrast, \methodName maintains highly dynamic scenes with stable subject and background appearance across all tested durations, despite relying solely on the pretrained Self-Forcing model, which natively supports only 5-second generation at 16\,FPS. These results highlight the robustness and scalability of our method for long-form autoregressive video generation.

\subsection{Discussion on Action Control Results}
Beyond autoregressive baselines, we also compare \methodName with LongLive~\cite{yang2025longlive}, SkyReels-V2~\cite{chen2025skyreels}, and Self-Forcing~\cite{huang2025self} for action-controlled video generation. LongLive introduces \emph{KV-Recache}, a cache management mechanism designed to enable prompt-dependent action transitions in autoregressive models. At each transition point, KV-Recache extracts all cached tokens, applies cross-attention with the new prompt, rebuilds a modified cache, and conditions subsequent frames on these recached latent tokens. This procedure aims to overwrite residual semantics from the previous prompt and improve responsiveness to new user instructions. However, the approach presents two main limitations:

\begin{enumerate}
    \item \textbf{Incomplete removal of previous prompt content.} During long rollouts, embeddings from earlier prompts accumulate in the cache and are not fully erased. This reduces prompt responsiveness over time and increases the delay before the new action appears. As demonstrated in the \textbf{Action Control Comparison} section of the project page, LongLive shows reduced responsiveness and significant identity and background drift, while \methodName preserves subject identity and background stability and responds to new prompts immediately.

    \item \textbf{Additional latency proportional to cache size and number of transitions.} KV-Recache requires reconstructing the entire cache at every prompt change. The resulting overhead increases with the cache size and the number of action transitions. By contrast, \methodName achieves instant prompt responsiveness by simply flushing the stale cache content, which is implemented as a local update to the cache end index without proportional computation.
\end{enumerate}

\subsection{Discussion on Dynamic Scene Cut Results}
Autoregressive video diffusion models naturally produce temporally smooth sequences, which often results in limited scene dynamism. To introduce cinematic variation without compromising coherence, we propose \text{RoPE Cut}, a mechanism that applies controlled discontinuities to the temporal RoPE coordinates. This technique enables intentional scene shifts while preserving overall generative stability.

We present our Dynamic Scene Cut results in the \textbf{project page} and in Fig.~\ref{fig:appendix_scene_cut}. Using RoPE Cut, we generate trailer-style sequences for several films, including \text{Harry Potter}, \text{Titanic}, \text{Game of Thrones}, \text{The Shawshank Redemption}, \text{Barbie}, and \text{Interstellar}. As demonstrated by these results, RoPE Cut produces dynamic, diverse scenes with varying backgrounds and environments within a single continuous generation stream, while consistently maintaining subject identity and visual fidelity.

\section{Interpretability via Attention Maps}
\label{sec:appendix_attention_maps}

\paragraph{Construction of frame-level attention maps.}
For each video, we extract self-attention weights from the middle transformer block of the denoiser at a fixed denoising step and aggregate them at the frame level. Let the video consist of $T$ frames, and let each frame $t$ be represented by a set of latent tokens. Denote by $a_{(t,i)\rightarrow (s,j)}$ the self-attention weight from query token $(t,i)$ (frame $t$, spatial index $i$) to key token $(s,j)$ (frame $s$, spatial index $j$), averaged over attention heads. We then construct a $T \times T$ frame–frame attention matrix $M$ by summing over all token pairs between frames:
\[
M_{t,s} = \sum_{i \in \text{frame } t} \sum_{j \in \text{frame } s} a_{(t,i)\rightarrow (s,j)}.
\]
Each cell $(t,s)$ in the attention map therefore corresponds to the total attention mass from all tokens of frame $t$ (query frame) to all tokens of frame $s$ (key frame). Since the underlying self-attention weights are row-normalized by the softmax, each row of $M$ is naturally normalized as well and can be interpreted as a frame-level attention distribution over the video history. A sharp diagonal structure means that each frame mainly attends to itself and its immediate temporal neighbors, while vertical stripes or off-diagonal blocks indicate longer-range dependencies or special tokens (e.g., sink tokens).

\paragraph{Block-Relativistic RoPE for Infinite-length Video Generation.}
For standard infinite-length generation, the \text{Block-Relativistic RoPE} map (\cref{fig:attn_map_1}) exhibits two main structures:
(i) a sharp diagonal band around the main diagonal, and
(ii) a persistent bright column corresponding to the global attention sink token.
The diagonal band shows that each query frame primarily attends to a small window of its recent predecessors and itself, which captures local temporal continuity and smooth motion. The bright sink column indicates that the model also consistently attends to a global sink token that provides a stable global context over time.

Crucially, all tokens that lie within the active KV window share a consistent local RoPE coordinate frame: their relative temporal indices stay within the range seen during pretraining (the teacher horizon), even though the absolute video length keeps growing. The attention maps do not show any drift of attention towards extremely early frames or degenerate patterns as the sequence becomes long. This supports our claim that Block-Relativistic RoPE re-anchors temporal indices within the teacher horizon instead of letting them drift to unseen absolute positions, effectively bypassing the 1024-index limit and enabling continuous, stable infinite-horizon rollouts.

\paragraph{KV Flush for Action-controllable Long Video Generation.}
For action-controllable generation, the \text{KV Flush} map (\cref{fig:attn_map_2}) visualizes the effect of our selective cache renewal strategy when the prompt is changed mid-generation. At the moment of a prompt change, we flush the KV cache and retain only two anchors: the global attention sink token and the last few generated frames (e.g., the last one or two frames before the change). All earlier frames are removed from the cache.

In the attention map, this appears as:
(i) a strong vertical column corresponding to the sink token, and
(ii) a narrow band of attention centered around the last pre-flush frame(s),
while attention to older frames is strongly suppressed and appears almost dark. After the flush, new frames attend primarily to the sink and these recent anchors, rather than to the distant history. This pattern confirms that the model re-orients its temporal context around a very short window of frames while keeping a stable global reference via the sink token.

Mechanistically, this behavior matches the intended design: the model preserves immediate temporal continuity (short-term motion and appearance consistency) through the last cached frames, but rapidly adapts to the new semantic guidance specified by the updated text prompt. The attention maps therefore make explicit how KV Flush balances short-term temporal coherence with fast semantic re-steering.

\paragraph{RoPE Cut for Dynamic Scene Transitions.}
For dynamic scene transitions, the \text{RoPE Cut} map (\cref{fig:attn_map_3}) shows what happens when we perform a scene cut. At the cut point, we apply two operations simultaneously:
(i) we flush the KV cache, and
(ii) we discontinuously advance the RoPE indices for the new frames, assigning them to a new temporal region that does not overlap with the pre-cut segment.

In the frame-level attention maps, this dual action produces two almost disjoint diagonal blocks. The first block corresponds to the pre-cut scene: frames in this block attend strongly to each other but receive very little attention from the post-cut frames. The second block corresponds to the post-cut scene: new frames attend mainly to themselves, their nearby neighbors, and the sink token, while giving only weak attention to frames from the first block. The weak residual attention to the pre-cut frames appears as a faint background, indicating that the old scene is still technically part of the extended context but is functionally de-emphasized.

This pattern shows that RoPE Cut forces the model to reset its effective temporal context at the cut: the post-cut segment behaves like a new scene with its own local temporal structure, rather than a continuation of the old one. At the same time, the sink-mediated pathway still provides a stable identity signal, which helps preserve subject identity across the scene boundary. In other words, the attention maps directly visualize how RoPE Cut implements a hard scene transition in the temporal representation, while still maintaining the global subject and style consistency learned by the model.

\end{document}